\pdfoutput=1

\documentclass[11pt]{article}

\usepackage[preprint]{acl}
\usepackage{times}
\usepackage{latexsym}

\usepackage{amsmath}
\usepackage{amssymb}
\usepackage{threeparttable}
\usepackage{multirow}
\usepackage{CJKutf8}
\usepackage{multirow}
\usepackage{graphicx}
\usepackage{booktabs}
\usepackage{threeparttable}

\usepackage[T1]{fontenc}

\usepackage[utf8]{inputenc}

\usepackage{microtype}

\usepackage{inconsolata}

\title{LAiW: A Chinese Legal Large Language Models Benchmark}

\author{Yongfu Dai$^{a}$, {\bf Duanyu Feng}$^{a}$, {\bf Jimin Huang}$^{b}$, {\bf Haochen Jia}$^{a}$, {\bf Qianqian Xie}$^{c}$,\\ {\bf Yifang Zhang}$^{a}$, {\bf Weiguang Han}$^{c}$, {\bf Wei Tian}$^{d}$, {\bf Hao Wang$^{a,}$\thanks{This is the corresponding author.}}\\
$^a$Sichuan University, $^{b}$ChanceFocus (Shanghai) AMC,\\ $^{c}$Wuhan University, $^{d}$Southwest Petroleum University\\
\texttt{wal.daishen@gmail.com}\quad  \texttt{fengduanyu@stu.scu.edu.cn}\quad \texttt{jimin@chancefocus.com}\\
\texttt{wwx990211@gmail.com}\quad \texttt{xqq.sincere@gmail.com}\quad \texttt{zhangyf\_ivy@foxmail.com}\\
\texttt{han.wei.guang@whu.edu.cn}\quad \texttt{418818347@qq.com}\quad \texttt{wangh@scu.edu.cn}
}

\begin{document}
\maketitle

\begin{abstract}

General and legal domain LLMs have demonstrated strong performance in various tasks of LegalAI. However, the current evaluations of these LLMs in LegalAI are defined by the experts of computer science, lacking consistency with the logic of legal practice, making it difficult to judge their practical capabilities. To address this challenge, we are the first to build the Chinese legal LLMs benchmark LAiW, based on the logic of legal practice. To align with the thinking process of legal experts and legal practice (syllogism), we divide the legal capabilities of LLMs from easy to difficult into three levels: basic information retrieval, legal foundation inference, and complex legal application. Each level contains multiple tasks to ensure a comprehensive evaluation. Through automated evaluation of current general and legal domain LLMs on our benchmark, we indicate that these LLMs may not align with the logic of legal practice. LLMs seem to be able to directly acquire complex legal application capabilities but perform poorly in some basic tasks, which may pose obstacles to their practical application and acceptance by legal experts. To further confirm the complex legal application capabilities of current LLMs in legal application scenarios, we also incorporate human evaluation with legal experts. The results indicate that while LLMs may demonstrate strong performance, they still require reinforcement of legal logic.
\end{abstract}

\section{Introduction}

With the emergence of ChatGPT\footnote{\url{https://openai.com/blog/chatgpt}} and GPT-4\footnote{\url{https://openai.com/gpt-4}} and their excellent text processing capabilities \cite{zhao2023survey}, many researchers have paid attention to the applications of large language models (LLMs) in various fields \cite{wang2023huatuo, xie2023pixiu, ko2023can}. In the field of legal artificial intelligence (LegalAI), which specifically studies how artificial intelligence can assist in legal construction \cite{zhong2020does, locke2022case, feng2022legal}, LLMs, especially those specializing in Chinese law, also show strong capabilities in generating legal text \cite{cui2023chatlaw,LAWGPT,HanFei}.

Chinese legal LLMs currently cover a wide range of legal tasks and undergo training through two different methods. 
While LLMs like HanFei \cite{HanFei} are pre-trained with large-scale legal documents, most Chinese legal LLMs, such as Fuzi-Mingcha \cite{fuzi.mingcha}, LexiLaw\footnote{\url{https://github.com/CSHaitao/LexiLaw}}, and ChatLaw \cite{cui2023chatlaw}, are fine-tuned with specific legal tasks based on the LLMs. The legal tasks they focus on have been well-defined in the field of NLP and can directly reflect their application effectiveness \cite{zhong2020does, choi2023use, steenhuis2023weaving}, such as legal question answering and consultation. Existing benchmarks for evaluating these models are also constructed based on these tasks \cite{yue2023disclawllm,fei2023lawbench}. However, due to the feasibility, these tasks are ultimately defined by computer experts, which may lack the training data and evaluation tasks for the logic of legal practice. Whether LLMs can currently be accepted and used by legal experts in practice is worth considering.

The application of LLMs in law should adhere to the logic of legal practice, known as the legal syllogism, involving the acquisition of evidence, legal articles, conclusions, and their interconnections \cite{kuppa2023chain, trozze2023large}. We expect legal LLMs to become true legal experts, and their handling of legal issues should involve three stages. Firstly, the ability to extract information from legal texts, then the ability to provide a reliable and reasoned answer based on solid legal knowledge, and ultimately the ability to form a complete response. For example, in criminal law, when judging someone, we need to first find relevant legal articles based on evidence, then calculate the judgment result based on these articles, and provide a well-organized and logical judgment text. 
More importantly, for courtroom officials, whether are LLMs capable of the above-mentioned abilities is their primary concern, which ensures the fairness of the law\footnote{\url{https://github.com/liuchengyuan123/LegalLLMEvaluation/}}.

In this work, to investigate the above-mentioned issue, we propose the first Chinese legal LLM benchmark LAiW\footnote{It means "AI in LAW".} based on the logic of legal practice. In this benchmark, following the guideline of the logic of legal practice, we categorize the legal capabilities of LLMs into three levels, from simple to difficult: basic information retrieval (BIR), legal foundation inference (LFI), and complex legal application (CLA). Among them, BIR focuses on the general NLP capabilities of LLMs and some legal evidence, knowledge, and category determination; 
LFI emphasizes the performance of LLMs in simple application tasks in the legal domain, which are typically components of more complex legal issues and interim results in legal logic; 
CLA focuses on the performance of LLMs in complex tasks in the legal domain, which require support from the abilities developed in the first two levels and some complex logical reasoning capabilities.
Based on these capabilities, our benchmark contains 14 tasks, covering most of the existing LegalAI tasks, and some new tasks. 

When conducting benchmark evaluations, we performed both automated evaluations and additional manual evaluations. For automated evaluations, we not only evaluate existing Chinese legal LLMs such as Lawyer-LLaMA \cite{lawyer-llama-report}, ChatLaw\cite{cui2023chatlaw} and HanFei \cite{HanFei}, but also focused on the base models of these Chinese legal LLMs and more effective general LLMs. 
The results of automated evaluations indicate that while existing LLMs have strong text generation capabilities for complex legal applications, they are unable to meet the underlying logic in legal applications in basic information retrieval and legal foundation inference. 
This may lead to significant distrust among legal experts when using LLMs. 
Therefore, we conduct additional manual evaluations specifically to assess the reliability of LLMs in complex legal applications. 
Through evaluations by legal experts, we find that in some complex legal applications with relatively lenient requirements for legal logic, LLMs' powerful generation ability cleverly bridges the gap in legal logic. 
However, in more demanding scenarios, they still exhibit significant discrepancies from real results. 
This further indicates the need for the tasks to train and evaluate the legal logic capabilities of LLMs.

Our contributions are as follows:
\begin{itemize}
    \item 
    We are proud to introduce the first Chinese legal LLMs benchmark LAiW, which is designed based on the logic of legal practice. We categorize the legal capabilities of LLMs into three levels: basic information retrieval, legal foundation inference, and complex legal application. This detailed categorization enables a more accurate assessment of the true abilities of LLMs in legal practice.
    \item Based on our automated evaluation, we demonstrate that the current capabilities of legal LLMs do not align with the demands of real-world applications. While LLMs demonstrate strong text generation abilities to complete complex legal applications, they struggle to achieve satisfactory performance in foundational legal logic tasks, making it difficult to gain the trust of legal experts.
    \item To assess whether the outstanding performance of LLMs in our third level is applicable in real scenarios, we invite legal experts for manual evaluations. Despite the LLMs' strong text generation capability, they still expose limitations with legal logic in practical applications. Therefore, We further advocate for the construction of training tasks to better reflect legal logic.
\end{itemize}

\section{Related Work}

\textbf{Chinese Legal LLMs.} 
We summarize the current Chinese legal LLMs and some general models in Table \ref{tab:LAiW-LLMs}.
Most of these Chinese legal LLMs focus on the ultimate application tasks in the legal field and are generally fine-tuned on some general LLMs.
For instance, LawGPT\_zh \cite{LAWGPT-zh}, Lawyer-LLaMA \cite{lawyer-llama-report}, ChatLaw \cite{cui2023chatlaw}, Fuzi-Mingcha \cite{fuzi.mingcha}, and LexiLaw developed the ability to answer legal questions and provide legal consultations by fine-tuning on related legal data. 
To compensate for the lack of legal knowledge due to only fine-tuning, these LLMs introduce additional legal knowledge databases for retrieval to supplement.
However, the accuracy and comprehensiveness of the knowledge base may be a major limiting factor for these LLMs.
The other Chinese legal LLMs adopted the pretraining or continued pretraining to enhance the legal knowledge of LLMs, such as LaWGPT \cite{LAWGPT}, wisdomInterrogatory\footnote{\url{https://github.com/zhihaiLLM/wisdomInterrogatory}}, and HanFei \cite{HanFei}.
They collect a large amount of legal text data, covering a wider range of legal tasks such as element extraction and case classification. These have a noticeable impact on improving the overall effectiveness of LLMs in legal applications.
However, the Chinese legal LLMs mentioned above mainly focus on the outcome of legal application (such as similarity to standard answers and fluency of generated text), which rarely consider whether they meet the logical requirements of legal practice. It is important to evaluate their legal logic, which is of utmost concern to legal experts.

\textbf{Legal LLMs Benchmark.} 
The development of LegalAI has led to a large number of tasks that combine law and computer science, from NLP-focused legal NER\footnote{\url{https://github.com/china-ai-law-challenge/CAIL2021/tree/main/xxcq}} and legal text summarization \cite{kanapala2019text} to legal-focused similar case matching \cite{locke2022case, sansone2022legal}, providing ample data for evaluating Chinese legal LLMs \cite{zhong2020does}. 
When categorizing from a legal perspective, it also encompasses the logic of the entire legal process from the legal elements extraction \cite{cao2022cailie, zhang2022recognition, zhong2020iteratively} to legal judgment prediction \cite{feng2022legal, cui2023survey}. 
Based on these tasks, LawBench \cite{fei2023lawbench} built an automatic evaluation framework for Chinese legal LLMs, which concerns the memorization, understanding, and application of legal knowledge.
DISC-Law-Eval Benchmark \cite{yue2023disclawllm} also based on the aforementioned tasks divides the evaluation into objective and subjective parts. The objective section assesses knowledge retention and reasoning abilities in the legal examination, and the subjective part uses GPT-3.5 Turbo to score the accuracy, completeness, and clarity of the answers. These frameworks greatly helped us understand the capabilities and improvement directions of current legal LLMs from the perspective of computer experts, especially the knowledge systems. However, whether these LLMs can be accepted by legal experts is also a question worthy of evaluation. In this work, we focus on addressing this issue from the logic of legal practice.



\begin{table*}[!htb]
  \centering
  \resizebox{0.999\textwidth}{!}{
  \begin{threeparttable}
  \begin{tabular}{lcccccc}
    \toprule
    Capability & Task & Primary Origin Dataset   & LAiW & Domain & Task Type\\
    \midrule
    
    \multirow{5}[0]{*}{BIR}  & Legal Article Recommendation & CAIL2018 \cite{xiao2018cail2018}  &  1000  & Criminal & Classification \\
    \cmidrule{2-6}  & Element Recognition & CAIL-2019 \cite{zhang2022recognition}  & 1000 & Civil & Classification \\
    \cmidrule{2-6}  & Named Entity Recognition & CAIL-2021 \cite{cao2022cailie} & 1040 & Criminal & Named Entity Recognition \\
    \cmidrule{2-6}  & Judicial Summarization & CAIL-2020 \cite{huang2023high} & 364 & Civil &  Text Generation \\
    \cmidrule{2-6}  & Case Recognition & CJRC \cite{duan2019cjrc} & 2000 & Criminal, Civil & Classification \\
    
    \midrule
    \multirow{6}[0]{*}{LFI} & Controversy Focus Mining & LAIC-2021  & 306 & - & Classification \\
    \cmidrule{2-6}  & Similar Case Matching & CAIL-2019 \cite{xiao2019cail2019}  & 260 & Civil & Classification \\
    \cmidrule{2-6} & Charge Prediction & Criminal-S \cite{hu2018few}  & 827 & Criminal & Classification \\
    \cmidrule{2-6} & Prison Term Prediction & MLMN \cite{ge2021learning}  & 349 & Criminal & Classification \\
    \cmidrule{2-6}  & Civil Trial Prediction & MSJudeg \cite{ma2021legal} & 800 & Civil & Classification \\
    \cmidrule{2-6}  & Legal Question Answering & JEC-QA \cite{zhong2020jec} & 855 & -  & Classification \\

    \midrule
    \multirow{3}[0]{*}{CLA} & Judicial Reasoning Generation & AC-NLG \cite{wu2020biased} & 834 &  Civil & Text Generation \\
    \cmidrule{2-6}  & Case Understanding & CJRC \cite{duan2019cjrc} & 1054 & Criminal, Civil & Text Generation \\
    \cmidrule{2-6}  & Legal Consultation & CrimeKgAssitant \cite{LAWGPT-zh} & 916 & - & Text Generation \\
    
    \bottomrule
    \end{tabular}

    \end{threeparttable}
}
  \centering
  \caption{Statistical information of our dataset. All datasets are sourced from open-source.}
  \label{tab:dataset}%

\end{table*}%

\section{Benchmark Construction}

In this section, we divide LLMs' abilities levels based on The Logic of Legal practice and construct our Chinese legal LLMs benchmark LAiW based on these levels. To ensure comprehensive evaluation, we cover both automatic evaluation using computable metrics and manual evaluation by legal experts.

\begin{figure}[!htb]
\centering
\includegraphics[width=1\columnwidth]{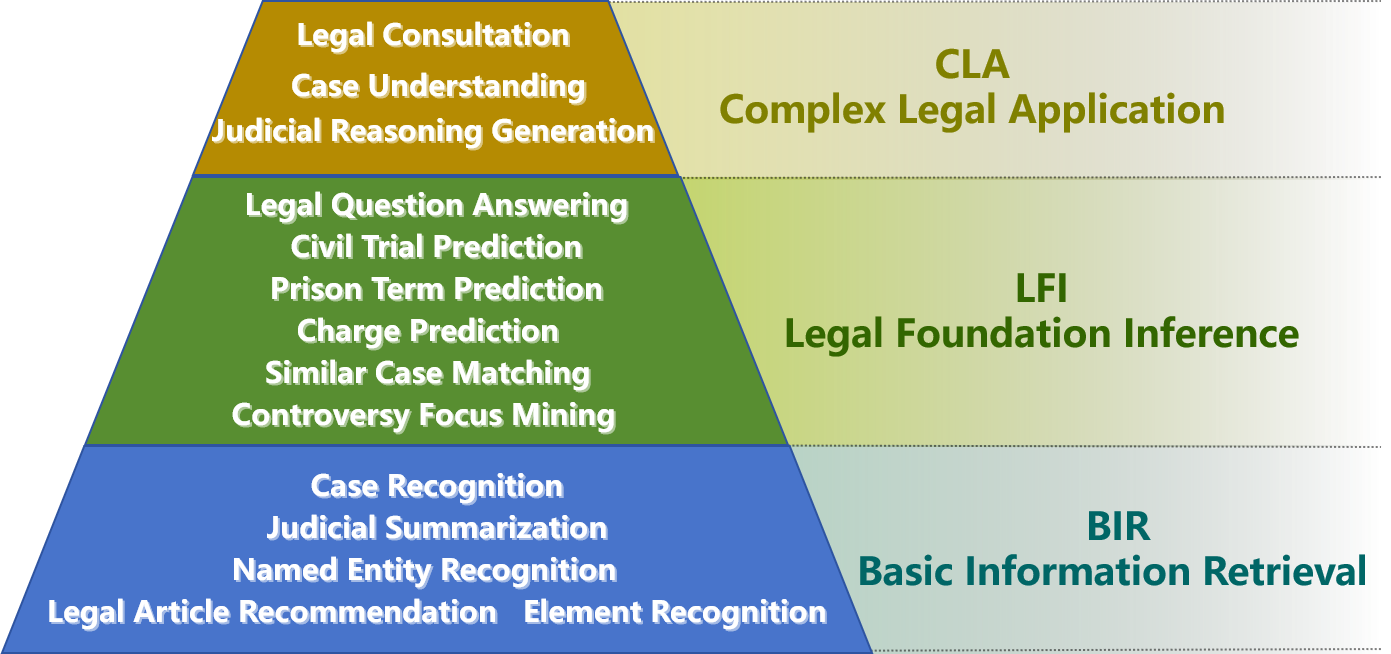} 
\caption{Multi-level Legal Capabilities of LLMs.}
\label{task_framework_en}
\end{figure}

\subsection{ The Logic of Legal Practice for LLMs} 
In contemporary legal practice, the logic is primarily based on Syllogism \cite{wroblewski1974legal, patterson2013logic}. Syllogism typically consists of three parts: the major premise, the minor premise, and the conclusion which is derived from the major and minor premises. In legal practice, this entails assessing the information and evidence pertinent to a case (minor premise), identifying the relevant legal articles (major premise), and reaching a judicial decision based on these factors (conclusion).

To ensure that LLMs also have the aforementioned logical thinking and remain synchronized with legal practice, we should also divide the abilities of LLMs into the aforementioned logical stages with 14 tasks. 
Specifically, we categorize the legal abilities of LLMs into three levels and try to align them with the logic of legal practice, shown in Figure \ref{task_framework_en}. 
By merging the process of acquiring minor premise and major premise, we construct the capability level of \textbf{basic information retrieval}. Building upon this foundation, we develop the capability level of \textbf{legal foundation inference} to draw preliminary conclusions based on the minor and major premises. Additionally, to assess the direct representation of the entire legal syllogism, we have created the capability level of \textbf{complex legal application}\footnote{Appendix \ref{appendix:dataset auto} provides detailed explanations for tasks of each capability level.}.

\subsubsection{BIR: Basic Information Retrieval}
We design the Basic Information Retrieval level with 5 tasks to assess the fundamental abilities of LLMs in legal logic, corresponding to directly accessible text information, minor premises, and major premises, such as legal evidence, legal knowledge, and category determination.

Specifically, we first consider three tasks that are well-defined in the fields of law and NLP: Named Entity Recognition, Judicial Summarization, and Case Recognition. 
They identify and summarize the key elements in legal texts, and classify cases as either Criminal or Civil.
Although these tasks may not require extensive legal knowledge from LLMs, they can yield a wealth of foundational information useful for both legal and computational purposes from the text.

We also consider two other tasks in the legal domain, namely Legal Article Recommendation and Element Recognition. 
The first task is to catch the major premises by finding relevant legal articles. The second task is to catch minor premises by identifying their relevant elements.

\subsubsection{LFI: Legal Foundation Inference} 
The Legal Foundation Inference level follows Syllogism's idea to explore the ability of LLMs to derive basic results and some judgment conclusions from minor premises and major premises.

We can divide 6 tasks for this capability into three parts. 
The first part presents the basic results of some simple legal applications, including Controversial Focus Mining and Similar Case Matching. 
Controversial Focus Mining is an intermediate result obtained in civil law based on the underlying circumstances and legal articles, used to determine the core issues of concern for both the plaintiff and the defendant.
Similar Case Matching involves finding similar cases based on the current case situation and referring to these cases to ensure the fairness of the judgment.
The second part involves predicting the outcomes of the court judgment conclusion. Since criminal law and civil law are two main branches of law, we have 3 tasks. Charge Prediction and Prison Term Prediction for criminal law, Civil Trial Prediction for civil law. 
Finally, The third part involves another application task, Legal Question Answering, that requires some fundamental integrated capabilities and focuses on the simple application of legal knowledge. 
Based on the information provided, LLMs provide some basic legal responses.

\subsubsection{CLA: Complex Legal Application} 
For this capability, we consider 3 challenging tasks that LLMs may be required to complete a complex legal reasoning and application task according to the entire logical process mentioned above.
Therefore, we focus on three tasks: Judicial Reasoning Generation, Case Understanding, and Legal Consultation.
Judicial Reasoning Generation involves the complete reproduction of the logical process from major and minor premises to conclusions in legal judgments. 
Case Understanding, on the other hand, analyzes the logic from the perspective of understanding, from major and minor premises to conclusion. 
Legal Consultation utilizes this logic from the perspective of a legal professional to provide assistance.

\subsection{Datasets Construction}
\label{Tasks-Construct}
Based on the mentioned criteria for capability division and task preparation, we combined the majority of open-source datasets with a small amount of proprietary data to construct the evaluation dataset for our LAiW benchmark. The dataset is divided into two parts: Automatic and Manual.

\subsubsection{Automatic Evaluation Legal Tasks}
To facilitate a more efficient evaluation of LLMs, we construct all 14 tasks mentioned above into datasets that can be automatically assessed shown in Table \ref{tab:dataset}.
The primary sources of this data include previous years' CAIL competition data \cite{xiao2018cail2018, zhang2022recognition, huang2023high}, as well as the most commonly used open-source data \cite{ge2021learning, wu2020biased, LAWGPT-zh} for these tasks, including various types of law, such as criminal law, civil law, constitutional law, social law, and economic law.

As for the task type of our dataset, it contains three types of tasks: classification, named entity recognition, and text generation. 
The classification refers to tasks where LLMs provide answers in the form of multiple-choice questions. 
Named entity recognition involves accurately identifying legal entities in the text, while text generation tasks allow LLMs to freely perform legal tasks and flexibly generate responses. 

During the construction of the dataset, we designed different prompts for various tasks to support LLMs in providing better answers. We initially validated the quality of prompts using ChatGPT and further confirmed their validity through legal experts. Currently, all tasks exist in a zero-shot format\footnote{Examples 
and the detailed processing methods can be found in Appendix \ref{appendix:dataset auto} and Appendix \ref{sec:Example Prompt}.}.

\subsubsection{Manual Evaluation Legal Tasks}
As shown in automatic evaluation results \ref{sec:automatic evaluation results}, we observed that these LLMs may not align with the logic of legal practice. LLMs seem to be able to directly acquire complex legal application capabilities but perform poorly in some basic tasks. To further confirm the complex legal application capabilities of current LLMs in legal application scenarios, we add a manual evaluation focus on the ability of the third level.

Due to the cost of manual evaluation, we focus on two tasks that are more oriented toward LLMs for logical reasoning: Judicial Reasoning Generation and Legal Consultation, rather than the Case Understanding task, which is based on the analysis of legal texts with existing logic \footnote{The detailed processing methods for the datasets are outlined in Appendix \ref{appendix:dataset manual}.}.

\begin{table*}[!htb]

  \centering
  
  \resizebox{0.92\textwidth}{!}{
  \begin{threeparttable}
  \begin{tabular}{ccccccc}
    \toprule
    Model & Model Size & Model Domain  & From & Baseline & Creater & Link \\
    \midrule

    GPT-4 \cite{openai2023gpt4} & - & General & Api & - & OpenAI & - \\
    \cmidrule{1-7}
    ChatGPT & - & General & Api & - & OpenAI & - \\
    \cmidrule{1-7}
    Baichuan2-Chat \cite{baichuan2023baichuan2} & 13B & General & Open & - & Baichuan Inc & \href{https://github.com/baichuan-inc/Baichuan2}{[1]} \\
    \cmidrule{1-7}
     Baichuan & 7B  & General  &  Open  &  - &  Baichuan Inc &  \href{https://github.com/baichuan-inc/Baichuan-7B}{[2]}  \\
     \cmidrule{1-7}
     ChatGLM \cite{du2022glm} & 6B  & General  & Open  &  - & Tsinghua, Zhipu  &  \href{https://github.com/THUDM/ChatGLM-6B}{[3]}  \\
     \cmidrule{1-7}
     Llama \cite{touvron2023llama} &  7B & General  &  Application &  - &  Meta AI &  \href{https://github.com/facebookresearch/llama}{[4]}  \\
     \cmidrule{1-7}
     Llama \cite{touvron2023llama} & 13B  & General  & Application  & -  & Meta AI  &  \href{https://github.com/facebookresearch/llama}{[4]}  \\
     \cmidrule{1-7}
     Llama2-Chat \cite{touvron2023llama2} & 7B & General & Application & - & Meta AI & \href{https://huggingface.co/meta-llama/Llama-2-7b-chat-hf}{[5]} \\
     \cmidrule{1-7}
     Chinese-LLaMA \cite{chinese-llama-alpaca}  & 7B  & General  &  Open & Llama-7B   &  Yiming Cui &  \href{https://github.com/ymcui/Chinese-LLaMA-Alpaca}{[6]}  \\
     \cmidrule{1-7}
     Chinese-LLaMA \cite{chinese-llama-alpaca}  & 13B  & General  &  Open & Llama-13B   &  Yiming Cui &  \href{https://github.com/ymcui/Chinese-LLaMA-Alpaca}{[6]}  \\
     \cmidrule{1-7}
     Ziya-LLaMA\cite{fengshenbang} &  13B &  General & Open  & Llama-13B &  IDEA-CCNL &  \href{https://huggingface.co/IDEA-CCNL/Ziya-LLaMA-13B-v1}{[7]}  \\
     \cmidrule{1-7}
     HanFei \cite{HanFei} & 7B  & Law  &  Open &  - & SIAT NLP  & \href{https://github.com/siat-nlp/HanFei}{[8]}   \\
     \cmidrule{1-7}
     wisdomInterrogatory  &  7B &  Law & Open  & Baichuan-7B  &  ZJU, Alibaba, e.t &  \href{https://github.com/zhihaiLLM/wisdomInterrogatory}{[9]}  \\
     \cmidrule{1-7}
     Fuzi-Mingcha \cite{fuzi.mingcha} & 6B  &  Law & Open  & ChatGLM-6B  & irlab-sdu  &  \href{https://github.com/irlab-sdu/fuzi.mingcha}{[10]} \\
     \cmidrule{1-7}
     LexiLaw & 6B  & Law  &  Open & ChatGLM-6B  &  Haitao Li &  \href{https://github.com/CSHaitao/LexiLaw}{[11]}  \\
     \cmidrule{1-7}
     LaWGPT \cite{LAWGPT} & 7B  &  Law & Open  & Chinese-LLaMA-7B  & Pengxiao Song  & \href{https://github.com/pengxiao-song/LaWGPT}{[12]}   \\
     \cmidrule{1-7}
     Lawyer-LLaMA \cite{lawyer-llama-report} &  13B & Law  & Open  &  Chinese-LLaMA-13B & Quzhe Huang  & \href{https://github.com/AndrewZhe/lawyer-llama}{[13]}   \\
     \cmidrule{1-7}
     ChatLaw \cite{cui2023chatlaw} &  13B & Law  &  Open & Ziya-LLaMA-13B  & PKU-YUAN's Group  & \href{https://github.com/PKU-YuanGroup/ChatLaw}{[14]} \\
    
    \bottomrule
    \end{tabular}
    \end{threeparttable}
}
\caption{The LLMs evaluated in our work. LaWGPT and wisdomInterrogatory undergo pre-training on Chinese-LLaMA and Baichuan respectively, followed by fine-tuning. HanFei does not have a baseline model. Apart from GPT-4 and ChatGPT, these general LLMs ave a parameter size of 7-13B to ensure a size similar to legal LLMs.}
  \label{tab:LAiW-LLMs}%
\end{table*}%

\section{Evaluation for Benchmark}
In this section, we provide our metrics and scoring method for automatic evaluation and our metrics and criteria for manual evaluation.

\subsection{Automatic Evaluation Legal Tasks}
\label{sec:Automatic Evaluation Legal Tasks}
Automatic Evaluation Legal Tasks contains classification tasks, named entity recognition tasks, and text generation tasks.
For classification tasks, we select accuracy (Acc), miss rate (Miss), F1 score (F1), and matthews correlation coefficient (Mcc) as evaluation metrics for these tasks. They can assess LLMs for their understanding of questions (Miss), the effectiveness of their answers (Acc and F1), and their ability to identify imbalances (Mcc). 
For named entity recognition tasks, we use the accuracy of the LLMs in identifying every legal entities (Entity-Acc). 
For text generation tasks, we use ROUGE-1 (R1), ROUGE-2 (R2), and ROUGE-L (RL) as evaluation metrics for this task. 

To evaluate the overall legal capabilities of LLMs, we further select a few key indicators for each task and calculate legal scores for LLMs based on these indicators as shown in Equation (\ref{eq:scores}).

\begin{equation}
\begin{cases}
    S_{\text{classification}}=F1*100, \\
    S_{\text{text generation}}=\frac{1}{3}(R1+R2+RL)*100, \\
    S_{\text{named entity recognition}}=\text{Entity-Acc}*100.
\end{cases}
\label{eq:scores}
\end{equation}

Then, the total score is calculated by averaging the scores of the three levels, which in turn are determined by averaging the scores of tasks within each level.

\subsection{Manual Evaluation Legal Tasks}
\label{sec:Manual Evaluation Legal Tasks}
First, to ensure the reliability of the assessment, we present criteria with several legal experts for manual evaluation. For the Judicial Reasoning Generation task, the criteria focus on completeness, relevance and accuracy. As for the Legal Consultation task, the criteria focus on fluency, relevance, and comprehensibility\footnote{A more detailed description is provided in Appendix \ref{Annotation rules and standards}.}.

We adopt the approach used in studies \cite{dubois2023alpacafarm,alpaca_eval} for manual evaluation, considering legal experts as evaluators, using reference answers as the baseline to calculate the win rate for the target LLMs. 
For example, when using the reference answer as the baseline, legal experts comprehensively assess the output of the target LLM and the reference answer from multiple judgment dimensions, and then choose the most satisfactory response.

\section{Experiment}

In this section, we present relevant experiment settings and highlight the key results.

\subsection{Experiment Settings}
\label{sec:auto evaluation}
For the automatic evaluation, We evaluate 18 LLMs, including 7 mainstream legal LLMs \cite{cui2023chatlaw, LAWGPT}, their corresponding 6 baseline LLMs \cite{du2022glm, chinese-llama-alpaca, zhang2022recognition}, and 5 more effective general LLMs \cite{baichuan2023baichuan2, touvron2023llama} such as GPT-4 and ChatGPT. For fairness in evaluation, all LLMs did not utilize legal knowledge databases. Table \ref{tab:LAiW-LLMs} lists more detailed information about these LLMs.

For the manual evaluation, We choose the top-performing four legal LLMs in our automatic evaluation. They are Fuz-Mingcha \cite{fuzi.mingcha}, HanFei \cite{HanFei}, Lawyer-LLaMa \cite{lawyer-llama-report}, and LexiLaw. Furthermore, we also conducted manual assessments of the performance of both GPT-4 and ChatGPT.

\begin{table*}[!htb]
  \centering

  \resizebox{0.999\textwidth}{!}{
  \begin{tabular}{ccccccccccccccccccc}
    \toprule
    \multirow{2}[2]{*}{Model} & \multicolumn{6}{c}{Basic Information Retrieva}  & \multicolumn{7}{c}{Legal Foundation Inference} & \multicolumn{4}{c}{Complex Legal Application} & \multirow{2}[2]{*}{Total Score} \\
    \cmidrule(lr){2-7}
    \cmidrule(lr){8-14}
    \cmidrule(lr){15-18}
     & $B_{1}$ & $B_{2}$ & $B_{3}$ & $B_{4}$ & $B_{5}$ & Avg. & $L_{1}$ & $L_{2}$ & $L_{3}$ & $L_{4}$ & $L_{5}$ & $L_{5}$ & Avg. & $C_{1}$ & $C_{2}$ & $C_{3}$ & Avg. & \\
     \midrule 
     
    GPT-4   & 99.20 & 82.27 & 80.67 & 42.72 & 99.75 & 80.92 & 80.50 & 45.94 & 100.00 & 65.58 & 70.43 & 53.14 & 69.27 & 37.22 & 96.19 & 42.66 & 58.69 & 69.63 \\
    ChatGPT  & 99.05 & 79.32 & 61.73 & 41.01 & 98.85 & 75.99 & 57.16 & 46.17 & 99.28 & 47.35 & 62.85 & 37.08 & 58.32 & 35.64 & 90.70 & 47.55 & 57.96 & 64.09 \\
    \midrule
    Baichuan2-13B-Chat & 45.07 & 52.18 & 47.31 & 26.67 & 97.14 & 53.67 & 4.12 & 2.99 & 17.50 & 61.43 & 67.91 & 38.24 & 32.03 & 52.61 & 81.29 & 41.31 & 58.40 & 48.04 \\
     Baichuan-7B   & 17.81 & 2.87 & 0.00 & 26.89 & 58.45 & 21.20 & 1.74 & 0.00 & 1.18 & 1.03 & 64.50 & 24.32 & 15.46 & 40.27 & 33.79 & 18.51 & 30.86 & 22.51 \\
     ChatGLM-6B    & 72.55 & 49.82 & 1.06 & 42.87 & 91.27 & 51.51 & 14.18 & 39.03 & 67.57 & 44.84 & 33.02 & 23.86 & 37.08 & 35.39 & 86.90 & 35.02 & 52.44 & 47.01 \\
     Llama-7B  & 19.53 & 1.43 & 0.00 & 11.40 & 23.23 & 11.12 & 1.31 & 0.00 & 35.19 & 1.03 & 49.15 & 5.74 & 15.40 & 0.61 & 56.08 & 10.93 & 22.54 & 16.35 \\
     Llama-13B   & 28.16 & 7.66 & 0.00 & 9.94 & 46.80 & 18.51 & 1.86 & 0.00 & 36.79 & 5.80 & 40.46 & 5.57 & 15.08 & 11.19 & 65.68 & 11.34 & 29.40 & 21.00 \\
     Llama2-7B-Chat   & 48.24 & 11.93 & 0.19 & 15.79 & 83.17 & 31.86 & 0.74 & 0.00 & 3.88 & 7.31 & 62.09 & 2.59 & 12.77 & 28.76 & 69.51 & 17.65 & 38.64 & 27.76 \\
     Chinese-LLaMA-7B & 24.39 & 7.45 & 0.00 & 30.77 & 48.97 & 22.32 & 2.02 & 0.76 & 31.79 & 1.03 & 65.24 & 8.63 & 18.25 & 26.34 & 62.31 & 13.81 & 34.16 & 24.91 \\
     Chinese-LLaMA-13B  & 30.34 & 5.47 & 0.00 & 7.73 & 61.56 & 21.02 & 3.28 & 5.05 & 20.21 & 5.33 & 64.46 & 16.60 & 19.16 & 18.86 & 73.15 & 12.40 & 34.80 & 24.99 \\
     Ziya-LLaMA-13B & 66.39 & 58.42 & 48.94 & 38.85 & 94.73 & 61.47 & 5.64 & 0.76 & 53.18 & 55.62 & 36.07 & 25.38 & 29.44 & 30.12 & 83.96 & 25.26 & 46.45 & 45.79 \\
    \midrule
     HanFei-7B & 24.91 & 7.25 & 51.63 & 21.14 & 82.18 & 37.42 & 1.15 & 0.00 & 5.27 & 2.73 & 66.81 & 22.03 & 16.33 & 51.31 & 81.19 & 27.43 & 53.31 & 35.69 \\
     wisdomInterrogatory-7B  & 0.39 & 0.19 & 0.00 & 34.75 & 27.99 & 12.66 & 3.57 & 35.38 & 2.32 & 1.30 & 16.76 & 3.34 & 10.45 & 13.91 & 68.02 & 18.17 & 33.37 & 18.83 \\
     Fuzi-Mingcha-6B & 58.95 & 12.58 & 0.38 & 47.92 & 78.57 & 39.68 & 4.70 & 20.84 & 31.53 & 48.40 & 32.66 & 26.64 & 27.46 & 49.55 & 80.48 & 34.10 & 54.71 & 40.62 \\
     LexiLaw-6B  & 47.16 & 2.89 & 31.35 & 41.79 & 83.43 & 41.32 & 2.11 & 18.49 & 3.40 & 6.42 & 4.35 & 18.51 & 8.88 & 25.85 & 80.81 & 24.52 & 43.73 & 31.31 \\
     LaWGPT-7B  & 10.15 & 2.59 & 0.00 & 27.69 & 36.92 & 15.47 & 1.62 & 0.00 & 20.04 & 1.03 & 54.55 & 8.40 & 14.27 & 35.23 & 65.62 & 14.11 & 38.32 & 22.69 \\
     Lawyer-LLaMA-13B & 20.26 & 1.52 & 7.88 & 51.13 & 73.44 & 30.85 & 2.19 & 0.76 & 0.24 & 2.12 & 12.75 & 20.26 & 6.39 & 34.00 & 85.68 & 31.83 & 50.50 & 29.25 \\
     ChatLaw-13B& 67.08 & 31.29 & 52.21 & 41.33 & 98.20 & 58.02 & 0.00 & 0.00 & 37.82 & 30.85 & 6.58 & 0.00 & 12.54 & 0.00 & 20.23 & 0.00 & 6.74 & 25.77 \\
     
    \bottomrule
    \end{tabular}
    }

    \caption{The scores of LLMs at various levels of the LAiW based on equation (\ref{eq:scores}). Here, $B_1$ to $B_5$ respectively represent the tasks: Legal Article Recommendation, Element Recognition, Named Entity Recognition, Judicial Summarization, and Case Recognition. $L_1$ to $L_6$ respectively represent the tasks: Controversy Focus Mining, Similar Case Matching, Charge Prediction, Prison Term Prediction, Civil Trial Prediction, and Legal Question Answering. $C_1$ to $C_3$ respectively represent the tasks: Judicial Reasoning Generation, Case Understanding, and Legal Consultation.}
    \label{tab:scores}%
\end{table*}%

\subsection{Automatic Evaluation Results}
\label{sec:automatic evaluation results}

The scores for each level and the total score of our automated evaluation are shown in Table \ref{tab:scores}\footnote{The complete results of each task are available in Appendix \ref{appendix:results of LLMs evaluation}.}.
We analyze the results from three different aspects: overall results, the legal logic of Chinese Legal LLMs, and the capabilities of Chinese Legal LLMs.

\textbf{Overall results.} When compared to GPT-4 and ChatGPT, there still exists a significant gap between the current open-source LLMs and specifically trained legal LLMs.

From Table \ref{tab:scores}, we find that GPT-4 and ChatGPT maintain optimal performance in most tasks. 
They significantly outperform the current open-source LLMs at various levels of scoring.
Among the open-source LLMs, only Baichuan2-Chat, ChatGLM, and Ziya-LLaMA achieve a total score of 45 or above. However, their performance in the BIR and LFI levels still lags far behind GPT-4 and ChatGPT.
As for the specifically trained legal LLMs, the top four performing ones are Fuzi-Mingcha, HanFei, LexiLaw, and Lawyer-LLaMA. However, their overall scores are lower, all below 45.

We believe that the reason for this phenomenon is twofold: first, due to the large number of parameters in GPT-4 and ChatGPT; second, we during the pretraining phase, GPT-4 and ChatGPT may have been exposed to a larger amount of data. 
Since the open-source LLMs we selected are primarily aimed at the Chinese community, the data they collect may be more limited compared to GPT-4 and ChatGPT. 
GPT-4 and ChatGPT cover a wide range of legal data in multiple languages. In this case, it is reasonable for them to have higher scores in the BIR and LFI levels which focus on the basic legal logic and legal knowledge.

\textbf{ The legal logic of Chinese Legal LLMs.} 
The Legal LLMs may not have the logic of legal practice. While they demonstrate strong text generation abilities to complete complex legal applications, they perform poorly in basic information retrieval and legal foundation inference tasks.

Observing Table \ref{tab:scores}, it is evident that the majority of legal LLMs score nearly 20 points higher in the application of direct logic (CLA level) compared to the scores in BIR and LFI levels. 
This is contrary to the logic typically found in law. 
It suggests that these LLMs seem to have learned the patterns of generating legal texts directly, but have not grasped the legal reasoning behind these patterns. 
As a result, LLMs are unable to effectively identify the major and minor premises in law and lack the ability to reason to a conclusion.
However, for the BIR level, ChatLaw stands out among legal LLMs.
It instead has a strong ability at the BIR level, which may stem from the outstanding performance of its base model Ziya - LLaMA at this level.

This raises concerns that current legal LLMs may not meet the expectations of legal experts, posing potential risks to the trust of LLMs in the legal domain.

\textbf{ The capabilities of Chinese Legal LLMs.} Fine-tuned legal LLMs have improved the normativity of legal text generation but may lose the legal logic. 
In addition, for legal LLMs, undergoing additional pre-training on legal text may be the path to achieving various legal capabilities and legal logic.

From table \ref{tab:scores}, legal LLMs such as Fuzi-Mingcha, WisdomInterrogatory, LaWGPT and Lawyer-LLaMA compared to their base models exhibit improvements in CLA level. 
Most of these legal LLMs improve the normativity of generated texts in legal text generation through instruction tuning. However, they may lead to a decrease in ability at the BIR and LFI levels, which indicates that instruction tuning may not enable LLMs to possess legal logic.

On the other hand, legal LLMs like HanFei, which focus more on pre-training, may indicate how Chinese Legal LLMs acquire ability and logic.
HanFei, based on an older LLM structure (Bloomz), underwent extensive pre-training on legal texts and demonstrated capabilities on par with subsequent legal LLMs.
Furthermore, GPT-4 and ChatGPT, models with extensive pre-training on large corpora, also showed excellent performance at the BIR and LFI levels. 
These findings indicate that developing legal reasoning and comprehensive abilities may require learning from a significant amount of pre-training text, rather than just fine-tuning. 
\subsection{Manual Evaluation Results}
\begin{table}[!htb]
  \centering

  \resizebox{0.49\textwidth}{!}{
  \begin{tabular}{c|ccc|ccc}
    \toprule
    
    \multirow{2}[0]{*}{Model} & \multicolumn{3}{c|}{Judicial Reasoning Generation}  & \multicolumn{3}{c}{Legal Consultation}  \\
    \cmidrule{2-7}
     & Total Score & Win Rate & Std & Total Score  & Win Rate & Std \\
    \midrule
    GPT-4 & 44.72 & 0.38 & 0.18 & \underline{43.97} & \textbf{0.85} & 0.15 \\
    \cmidrule{1-7}
    ChatGPT & 41.74 & 0.35 & 0.27 & \textbf{48.79} & \underline{0.79} & 0.12 \\
    \cmidrule{1-7}
    Fuzi-Mingcha & \textbf{63.58} & \textbf{0.65} & 0.35 & 35.22 & 0.51 & 0.19 \\
    \cmidrule{1-7}
    HanFei & \underline{60.13}  & \underline{0.59} & 0.26 & 27.06 & 0.33 & 0.06 \\
    \cmidrule{1-7}
    LexiLaw & 43.48 & 0.31 & 0.15 & 25.53 & 0.24 & 0.02 \\
    \cmidrule{1-7}
    Lawyer-LLaMA & 39.61 & 0.30 & 0.26 & 33.27  & 0.51 & 0.21 \\
        
    \bottomrule
    \end{tabular}
}
\caption{The average win rate (WR) of LLMs for the Judicial Reasoning Generation and Legal Consultation tasks. The total score represents the score obtained by LLMs through automatic evaluation on our benchmark.}
  \label{tab:win rate}%
\end{table}%

According to the criteria for expert evaluation in Section \ref{sec:Manual Evaluation Legal Tasks} and the calculated average win rate scores of legal experts shown in Table \ref{tab:win rate}\footnote{The complete results are available in Appendix \ref{appendix:win rate for each expert} and the experts' agreement are available in Appendix \ref{The agreement scores}.}.
Based on these results, we have two findings.

\textbf{The lack of legal logic in LLMs still exists in Complex Legal Applications.}
For the task of Judicial Reasoning Generation that requires a strong understanding of legal logic, even models with powerful text generation capabilities like GPT-4 and ChatGPT may have deficiencies in legal logic. 
As described in Section \ref{sec:Manual Evaluation Legal Tasks}, the Judicial Reasoning Generation task focuses on accuracy, such as the correct citation of legal articles and reasoning based on the citation. This directly connects to the basic logic of legal.
Therefore, most of the LLMs' win rates are much lower than 0.5, indicating that strong text generation capabilities cannot directly replace legal logic.

For tasks like Legal Consultation, there is a lower requirement for legal logic but a higher requirement for fluency. 
Therefore, during the manual evaluation, legal experts tend to prefer models with stronger language capabilities, which is the strength of GPT-4 and ChatGPT. This capability can also be learned by legal LLMs through instruction tuning. 
As a result, the final evaluation results of legal experts also reflect this, giving higher win rates to all LLMs, with most even surpassing the annotated answers.

\textbf{Manual evaluation and automatic evaluation share similarities, enhancing the reliability of our automatic evaluation.}
From Table \ref{tab:win rate}, we can observe that the results of manual evaluation and automated evaluation are similar. 
For instance, in both evaluation rounds, Fuzi-Mingcha and HanFei performed best in the Judicial Reasoning Generation task, while GPT-4 and ChatGPT excelled in the Legal Consultation task. 
In addition, despite its shortcomings as an automated evaluation metric in many cases, Rouge still demonstrates a certain level of capability when reflecting legal logic.
This indicates that our automatic evaluation can provide a reliable path for the legal logic assessment of legal LLMs and further reduce manual effort. 
Additionally, our assessment of legal logic is granular, and the degree of emphasis on legal logic in different scenarios can also be reflected by our automatic evaluation of different tasks.

\section{Conclusion}
This paper aims to construct a Chinese Legal LLMs benchmark based on the logic of legal practice. To match the process of syllogism in legal logic, the benchmark categorizes LLM legal capabilities into basic information retrieval, legal foundation inference, and complex legal application, and includes 14 tasks. During benchmark evaluations, automated and manual evaluations were conducted. Automated results showed that existing LLMs excel in text generation for complex legal applications but struggle with basic information retrieval and legal foundation inference, leading to a lack of legal logic and distrust among legal experts. Manual evaluations revealed that while LLMs can bridge the gap in legal logic in some application scenarios, they still exhibit significant discrepancies as legal experts. This highlights the urgent need for better training and evaluation methods for LLMs in the legal domain. Therefore, our benchmark, for the first time from a legal perspective, reveals the professional legal capabilities of existing legal LLMs, greatly facilitating the subsequent evaluation and development work of legal LLMs.

\newpage
\section{Limitations and Future Work}
\label{sec:limitations}
Due to the significant amount of work involved in using a computer-based approach to construct the logic of legal practice related tasks and assessments, we also have the following two limitations and areas for future work:

1) In the manual evaluation experiment, to save workload, only a portion of the data and LLMs are sampled and chosen for evaluation, rather than assessing all of them. In the future, we will collaborate more with legal experts to ensure a more comprehensive human assessment.

2) Most of the tasks are derived from publicly available legal data, which may not fully evaluate the logic of legal practice for LLMs. We will further develop additional tasks to refine the logic of legal practice at each stage.




\section{Ethics Statement}
Due to the sensitivity of the legal field, we have conducted a comprehensive review of the relevant data in this benchmark. The open-source datasets we used all have corresponding licenses. We have masked sensitive information, such as names, phone numbers, and IDs, and legal experts have conducted ethical evaluations.


\section*{Acknowledgements}

\bibliography{anthology,custom}

\appendix

\section{Data Construction}

\subsection{Automatic Evaluation Dataset}
\label{appendix:dataset auto}
In this section, we provide the construction details for the LAiW datasets of each task.

\subsubsection{BIR: Basic Information Retrieval}

\textbf{Legal Article Recommendation.}

\textbf{Definition}: Legal Article Recommendation aims to provide relevant articles based on the description of the case.

\textbf{Description}: It comes from the first stage data of the CAIL-2018\footnote{\url{http://cail.cipsc.org.cn/task_summit.html?raceID=1&cail_tag=2018}}, aimed at providing relevant legal articles based on case descriptions. We selected the top three legal articles with their corresponding charges, namely the crime of dangerous driving, theft, and intentional injury. The three charges correspond to Article 133, Article 264, and Article 234 of the Criminal Law of the People's Republic of China.

\textbf{Prompt}: "Based on the relevant description provided below, predict the applicable law article. The options are ('133', '264', '234'). Your answer must be one of these three articles. These articles represent the legal provisions in the Criminal Law of the People's Republic of China. Among them, Article '133' refers to 'Violating regulations on transportation management, resulting in a major accident causing serious injury, death, or significant loss of public or private property'. Article '264' refers to 'Stealing public or private property, or committing theft multiple times, burglary, armed theft, or pickpocketing'. Article '234' refers to 'Intentionally causing bodily harm to others'. Text:"

\begin{CJK*}{UTF8}{gbsn}"请根据下面给定的案件的相关描述预测其涉及的法条，可供选择的法条为('133', '264', '234')，回答只能是这三个法条中的一个。这三个法条代表《中华人民共和国刑法》中的法律条文，其中，法条'133'表示'违反交通运输管理法规，因而发生重大事故，致人重伤、死亡或者使公私财产遭受重大损失'，法条'264'表示'盗窃公私财物，或者多次盗窃、入户盗窃、携带凶器盗窃、扒窃'，法条'234'表示'故意伤害他人身体'。 文本:"\end{CJK*}

\textbf{Element Recognition.}

\textbf{Definition}: Element Recognition analyzes and assesses each sentence to identify the pivotal elements of the case. 

\textbf{Description}: It comes from the element recognition track\footnote{\url{https://github.com/china-ai-law-challenge/CAIL2019}} of the CAIL-2019, aiming to automatically extract key factual descriptions from case descriptions. The original dataset primarily involves marriage, labor disputes, and loan disputes. We selected the labor dispute dataset.

\textbf{Prompt}: "Based on the partial paragraphs of the arbitral awards in the field of labor disputes below, identify the elements involved. The selectable elements are ('LB1', 'LB2', 'LB3', 'LB4', 'LB5', 'LB6', 'LB7', 'LB8', 'LB9', 'LB10', 'LB11', 'LB12', 'LB13', 'LB14', 'LB15', 'LB16', 'LB17', 'LB18', 'LB19', 'LB20'). The options are as follows: 'LB1' represents 'termination of labor relations', 'LB2' represents 'payment of wages', 'LB3' represents 'payment of economic compensation', 'LB4' represents 'non-payment of full labor remuneration', 'LB5' represents 'existence of labor relations', 'LB6' represents 'no labor contract signed', 'LB7' represents 'labor contract signed', 'LB8' represents 'payment of overtime wages', 'LB9' represents 'payment of double wages compensation for unsigned labor contracts', 'LB10' represents 'payment of work-related injury compensation', 'LB11' represents 'not raised at the labor arbitration stage', 'LB12' represents 'non-payment of compensation for illegal termination of labor relations', 'LB13' represents 'economic layoffs', 'LB14' represents 'non-payment of bonuses', 'LB15' represents 'illegally collecting property from workers', 'LB16' represents 'specialized occupations', 'LB17' represents 'payment of work-related death allowance|funeral allowance|bereavement allowance', 'LB18' represents 'advance notice of termination by the employer', 'LB19' represents 'corporate legal status has ceased', 'LB20' represents 'mediation agreement exists'. Text:"

\begin{CJK*}{UTF8}{gbsn}"请根据以下劳动争议领域的裁判文书的部分句段，识别其涉及的要素，可供选择的要素有('LB1', 'LB2', 'LB3', 'LB4', 'LB5', 'LB6', 'LB7', 'LB8', 'LB9', 'LB10','LB11', 'LB12', 'LB13', 'LB14', 'LB15', 'LB16', 'LB17', 'LB18', 'LB19', 'LB20')，回答只能是这二十个选项中的一个。这二十个选项中，'LB1'表示'解除劳动关系'，'LB2'表示'支付工资'，'LB3'表示'支付经济补偿金'，'LB4'表示'未支付足额劳动报酬'，'LB5'表示'存在劳动关系'，'LB6'表示'未签订劳动合同'，'LB7'表示'签订劳动合同'，'LB8'表示'支付加班工资'，'LB9'表示'支付未签订劳动合同二倍工资赔偿'，'LB10'表示'支付工伤赔偿'，'LB11'表示'劳动仲裁阶段未提起'，'LB12'表示'不支付违法解除劳动关系赔偿金'，'LB13'表示'经济性裁员'，'LB14'表示'不支付奖金'，'LB15'表示'违法向劳动者收取财物'，'LB16'表示'特殊工种'，'LB17'表示'支付工亡补助金|丧葬补助金|抚恤金'，'LB18'表示'用人单位提前通知解除'，'LB19'表示'法人资格已灭失'，'LB20'表示'有调解协议'。文本："\end{CJK*}

\textbf{Named Entity Recognition.}

\textbf{Definition}: Named Entity Recognition aims to extract nouns and phrases with legal characteristics from various legal documents.

\textbf{Description}: It comes from the Information Extraction competition of CAIL-2021, aiming to extract the main content of judgments\footnote{\url{https://github.com/isLouisHsu/CAIL2021-information-extraction/tree/master}}. The original dataset covers 10 legal entities, including "criminal suspect," "victim," etc. We selected five entities: "criminal suspect," "victim," "time," "stolen items," and "item value." We filtered out samples with non-nested entities. We used five prompts, each corresponding to one of the five legal entities.

\textbf{Prompt}: "Your task is to extract the entity 'suspect' from the text below. If this entity does not exist, the answer is 'No'. Text: "A set of stolen 'Jingqiu' brand batteries worth 1488 yuan." Answer:"

\begin{CJK*}{UTF8}{gbsn}"你的任务是从下面的文本中提取'犯罪嫌疑人'实体，如果不存在这个实体，则回答'No'。 文本：被盗“京球”牌蓄电池一组价值人民币1488元。 回答："\end{CJK*}

\textbf{Judicial Summarization.}

\textbf{Definition}: Judicial Summarization aims to condense, summarize, and synthesize the content of legal documents. 

\textbf{Description}: It comes from the Judicial Summary competition of CAIL-2020, aiming to extract the main content of judgments\footnote{\url{http://cail.cipsc.org.cn/task_summit.html?raceID=4&cail_tag=2022}}. We removed certain information from the original text of each sample, including case number, case title, judges, trial time, etc., as we believe this information has little impact on the quality of summary generation. Additionally, we only kept samples with a text length less than 1.5k.

\textbf{Prompt}: "Please extract an abstract from the legal document given below and express its main content in shorter, more coherent and natural words. text:"

\begin{CJK*}{UTF8}{gbsn}"请对下面给的这篇法律文书提取摘要，用更短、更连贯、更自然的文字表达其主要内容。 文本："\end{CJK*}

\textbf{Case Recognition.} 

\textbf{Definition}: Case Recognition aims to determine, based on the relevant description of the case, whether it pertains to a criminal or civil matter.

\textbf{Description}: It comes from CJRC, aiming to determine whether a given case is a criminal or civil case based on relevant case descriptions\footnote{\url{https://github.com/china-ai-law-challenge/CAIL2019/tree/master}}. We sampled criminal and civil cases in nearly a 1:1 ratio.

\textbf{Prompt}: "
Please determine whether the following case belongs to criminal or civil cases based on the title or relevant description text, and your response should be one of the two options. Text:"

\begin{CJK*}{UTF8}{gbsn}"请根据以下案件的标题或者相关描述文本，判断该案件属于刑事案件还是民事案件，并且你的回答应该只能是其中一个。 文本："\end{CJK*}

\subsubsection{LFI: Legal Foundation Inference}

\textbf{Controversy Focus Mining.} 

\textbf{Definition}: Controversial Focus Mining aims to extract the logical and interactive arguments between the defense and prosecution in legal documents, which will be analyzed as a key component for the tasks that relate to the case result.

\textbf{Description}: It comes from the Controversy Focus Recognition task of LAIC, aiming to identify and detect the disputed focal points based on the original plaintiff's claims and defense contents in legal judgments\footnote{\url{https://laic.cjbdi.com/}}. We selected samples that meet the following conditions: 1) contain only one disputed focal point, 2) have a text length less than 3k, and 3) involve the top ten disputed focal points in terms of frequency. Consequently, we restructured the dataset into a classification task, where the model is required to correctly identify the disputed focal point from the ten available options for each sample.

\textbf{Prompt}: "Please select the most appropriate dispute focus based on the plaintiff's claims and defendant's defense in the judgment document. The options are ('A', 'B', 'C', 'D', 'E', 'F', 'G', 'H', 'I', 'J'), representing ten dispute focuses respectively. You only need to return the letter of the correct option. Among them, 'A' represents 'determination of the amount of engineering funds', 'B' represents 'determination of the amount of damages compensation', 'C' represents 'dispute over principal/loan agreement/written agreement or electronic agreement/expressions of borrowing intention', 'D' represents 'dispute over principal/loan agreement/written agreement or electronic agreement/principal amount', 'E' represents 'liability determination', 'F' represents 'whether there is a breakdown of relationship', 'G' represents 'guarantee liability/claim for warranty', 'H' represents 'existence of labor relations', 'I' represents 'contractual effectiveness issue', 'J' represents 'responsibility assumption'. Text:"

\begin{CJK*}{UTF8}{gbsn}"请根据裁判文书中原被告的诉请及答辩内容选择出一个最匹配的争议焦点。可供选择的回答为('A', 'B', 'C', 'D', 'E', 'F', 'G', 'H', 'I', 'J')，这十个选项分别代表十个争议焦点，你只需要返回正确选项的字母。其中，回答'A'表示'工程款数额认定'，回答'B'表示'损失赔偿数额认定'，回答'C'表示'本金争议/借贷合意/书面协议or电子协议/借款的意思表示'，回答'D'表示'本金争议/借贷合意/书面协议or电子协议/本金（金额）'，回答'E'表示'责任认定'，回答'F'表示'感情是否破裂'，回答'G'表示'担保责任/保证责任诉求'，回答'H'表示'是否存在劳动关系'，回答'I'表示'合同效力问题'，回答'J'表示'责任承担'。 文本："\end{CJK*}

\textbf{Similar Case Matching.} 

\textbf{Definition}: Similar Case Matching aims to find cases that bear the closest resemblance, which is a core aspect of various legal systems worldwide, as they require consistent judgments for similar cases to ensure the fairness of the law.

\textbf{Description}: It comes from CAIL2019-SCM, which aims to match similar cases based on factual descriptions\footnote{\url{https://github.com/china-ai-law-challenge/CAIL2019/tree/master/scm}}. Each entry in the original dataset contains three fields labeled 'A,' 'B,' and 'C,' representing three legal factual descriptions. Our task is to determine, given three legal documents A, B, and C, which one (B or C) is more similar to A. Additionally, each selected case has a length not exceeding 2k.

\textbf{Prompt}: "Based on the content of Case A, select the case that is more similar to Case A. The options are ('B', 'C'). The length of the answer is limited to 3 characters, meaning you only need to provide the letter of the correct option. 'B' indicates that Case B is more similar to Case A, while 'C' indicates that Case C is more similar to Case A."

\begin{CJK*}{UTF8}{gbsn}"请根据案件A的内容，选择与案件A相似度更高的案件，可供选择的回答为（'B', 'C'），回答的文本长度限制为3个字符，即只回答正确选项的字母。其中，回答'B'表示案件B与案件A相似度更高，回答'C'表示案件C与案件A相似度更高。"\end{CJK*}

\textbf{Charge Prediction.}

\textbf{Definition}: It is the sub-task of Criminal Judgment Prediction task. Criminal Judgment Prediction involves predicting the guilt or innocence of the defendant, along with the potential sentencing, based on the results of basic legal NLP, including the facts of the case, the evidence presented, and the applicable law articles.

\textbf{Description}: It is from the Criminal-S dataset\footnote{\url{https://github.com/thunlp/attribute_charge}}, which consists of criminal cases published by CJO. As each case is well-structured and divided into multiple sections such as facts, court opinions, and judgment results, the authors of this dataset chose the facts section of each case as input and selected 149 different charges as output. In this paper, we specifically chose the charges of "Theft," "Intentional Smuggling," and "Drug Trafficking, Selling, Transporting, and Manufacturing" as our focus. Each sample corresponds to a unique charge.

\textbf{Prompt}: "Based on the given description of the case below, predict the crime it involves. The options are ('69', '50', '124'). You can only choose one of these three options. '69' represents 'theft', '50' represents 'intentional injury', and '124' represents 'smuggling, selling, transporting, or manufacturing drugs'. Text:"

\begin{CJK*}{UTF8}{gbsn}"请根据下面给定的案件的相关描述预测其涉及的罪名，可供选择的回答为('69', '50', '124')，回答只能是这三个选项中的一个。这三个选项代表了三个罪名，其中，罪名'69'表示'盗窃罪'，罪名'50'表示'故意伤害罪'，罪名'124'表示'走私、贩卖、运输、制造毒品罪'。 文本："\end{CJK*}

\textbf{Prison Term Prediction.}

\textbf{Definition}: It is the sub-task of Criminal Judgment Prediction task, which is defined in Charge Prediction task.

\textbf{Description}: It comes from MLMN, aiming to learn fine-grained correspondences of factual-Articles in legal cases\footnote{\url{https://github.com/gjdnju/MLMN}}. The original dataset is divided into crimes of injury and traffic accidents. Based on the original data's months of imprisonment, the labels are categorized into five classes. In this paper, we further categorized the sentences into three classes: the first class includes non-punishment and detention, the second class includes imprisonment of less than 1 year and 1 year to less than 3 years, and the third class includes imprisonment of 3 years to less than 10 years.

\textbf{Prompt}: "Based on the given description of the case below, predict the possible sentence the defendant may receive. The options are ('A', 'B', 'C'). You can only choose one of these three options. 'A' represents 'non-criminal punishment' or 'detention', 'B' represents 'fixed-term imprisonment of less than 3 years', and 'C' represents 'fixed-term imprisonment of 3 years or more but less than 10 years'. Text:"

\begin{CJK*}{UTF8}{gbsn}"请根据下面给定的案件的相关描述预测被告人可能被判的刑期，可供选择的回答为('A', 'B', 'C')，回答只能是这三个选项中的一个。这三个选项对应了三个刑期区间，其中，回答'A'表示'免予刑事处罚'或'拘役'，回答'B'表示'3年以下有期徒刑'，回答'C'表示'3年及3年以上，10年以下有期徒刑'。 文本："\end{CJK*}

\textbf{Civil Trial Prediction.}

\textbf{Definition}: Civil Trial Prediction task involves using factual descriptions to predict the judgment of the defendant in response to the plaintiff's claim, which we should consider the Controversial Focus.

\textbf{Description}: It comes from MSJudge, aiming to predict opinions on each claim based on case-related descriptions and claims\footnote{\url{https://github.com/mly-nlp/LJP-MSJudge}}. The original dataset includes court factual descriptions, multiple claims, and judgments for each claim. We extracted samples with only a unique claim and sampled them based on the distribution of judgment results.

\textbf{Prompt}: "Based on the factual description of the civil case provided below and a litigation request, provide an overall judgment prediction for the litigation request. Your response can only be one of the three options ('A', 'B', 'C'). 'A' indicates support for the litigation request, 'B' indicates partial support for the litigation request, and 'C' indicates opposition to the litigation request."

\begin{CJK*}{UTF8}{gbsn}"根据下面给定民事案件的事实描述和一个诉讼请求，给出你对该诉讼请求的一个整体裁判预测，你的回答只能是('A', 'B', 'C')三个选项中的一个。其中，'A'表示支持诉讼请求，'B'表示部分支持诉讼请求，'C'表示反对诉讼请求。"\end{CJK*}

\textbf{Legal Question Answering.}

\textbf{Definition}: Legal Question Answering utilizes the model's legal knowledge to address the national judicial examination, which encompasses various specific legal types.

\textbf{Description}: It is from a question-answering dataset collected from the China National Judicial Examination\footnote{\url{https://jecqa.thunlp.org/}}, which includes both single-choice and multiple-choice questions. The goal is to predict answers using the presented legal questions and relevant articles. We selected only the single-choice questions for our analysis.

\textbf{Prompt}: "Please answer the question based on the judicial examination question below. There is only one correct answer among the options ('A', 'B', 'C', 'D'). You don't need to provide a detailed analysis of the question, just select the correct answer."

\begin{CJK*}{UTF8}{gbsn}"请根据下面的司法考试题目回答问题，选项('A', 'B', ‘C', 'D')中只有一个正确答案。你不需要返回对题目的具体分析，只需选出正确的答案。"\end{CJK*}

\subsubsection{CLA: Complex Legal Application}

\textbf{Judicial Reasoning Generation.}

\textbf{Definition}: Judicial Reasoning Generation aims to generate relevant legal reasoning texts based on the factual description of the case. It is a complex reasoning task, because the court requires further elaboration on the reasoning behind the judgment based on the determination of the facts of the case. This task also involves aligning with the logical structure of syllogism in law.

\textbf{Description}: It comes from the AC-NLG dataset\footnote{\url{https://github.com/wuyiquan/AC-NLG}}, constructed from private lending cases, which are the most common category in civil cases. The focus is on the task of generating court opinions in civil cases. This task takes the plaintiff's claims and factual descriptions as input and generates the corresponding court opinions as output.

\textbf{Prompt}: "Please generate corresponding "the court holds that" content based on the "litigation requests" and "trial findings" provided in the brackets below."

\begin{CJK*}{UTF8}{gbsn}"请你根据下面中括号里的'诉讼请求'和'审理查明'内容生成对应的'本院认为'内容。"\end{CJK*}

\textbf{Case Understanding.}

\textbf{Definition}: Case Understanding is expected to provide reasonable and compliant answers based on the questions posed regarding the case-related descriptions in the judicial documents, which is also a complex reasoning task.

\textbf{Description}: It also comes from the CJRC dataset\footnote{\url{https://github.com/china-ai-law-challenge/CAIL2019/tree/master}}, which includes 10,000 documents and nearly 50,000 questions with answers. These documents are from judgment files, and the questions are annotated by legal experts. Each document contains multiple questions. In this paper, we selected only the training set from the original data, where each question has only one standard answer.

\textbf{Prompt}: "Based on the provided "legal text material" content, answer the corresponding "question" to complete the task of fragment extraction-based reading comprehension. Specifically, you need to correctly answer the "question", and the answer is limited to a clause (or fragment) from the "legal text material". Please provide your answer in the format '''Answer: A''', where A represents the correct clause (or fragment) from the "legal text material"."

\begin{CJK*}{UTF8}{gbsn}"请你根据下面提供的'法律文本材料'内容，回答相应的'问题'，以完成片段抽取式的阅读理解任务。 具体来说，你要正确回答'问题'，并且答案限定是'法律文本材料'的一个子句（或片段）。请你以'''答案：A'''的格式给出回答，其中A表示'法律文本材料'中正确的子句（或片段）。"\end{CJK*}

\textbf{Legal Consultation.}

\textbf{Definition}: Legal Consultation covers a wide range of legal areas and aims to provide accurate, clear, and reliable answers based on the legal questions provided by the different users. Therefore, it usually requires the sum of the aforementioned capabilities to provide professional and reliable analysis.

\textbf{Description}: It comes from the CrimeKgAssistant dataset\footnote{\url{https://github.com/LiuHC0428/LAW-GPT}}, where ChatGPT has been utilized to rephrase answers based on the Q\&A pairs from CrimeKgAssistant. The goal is to generate answers that are more detailed and linguistically well-organized compared to the original responses. We further filtered question-answer pairs by identifying responses containing phrases like \begin{CJK*}{UTF8}{gbsn}"抱歉"\end{CJK*} or \begin{CJK*}{UTF8}{gbsn}"无法准确回答"\end{CJK*}, and cases where questions contained numerous \begin{CJK*}{UTF8}{gbsn}"?"\end{CJK*} symbols or were linguistically awkward.

\textbf{Prompt}: "If you are a lawyer, please answer the legal consultation question below based on the real scenario."

\begin{CJK*}{UTF8}{gbsn}"假设你是一名律师，请回答下面这个真实情景下的中文法律咨询问题。"\end{CJK*}

\subsection{Some Automatic Evaluation Examples}
\label{sec:Example Prompt}

This is an appendix for examples of our three tasks: CLS, NER and TG.
For the CLS task, we choose the CR task. Then we respectively choose CR, NER, LC as example prompts for these three tasks.

In NER task, the \{entity\} include: \begin{CJK*}{UTF8}{gbsn}犯罪嫌疑人\end{CJK*},\begin{CJK*}{UTF8}{gbsn}受害人\end{CJK*}, \begin{CJK*}{UTF8}{gbsn}时间\end{CJK*}, \begin{CJK*}{UTF8}{gbsn}物品价值\end{CJK*}, \begin{CJK*}{UTF8}{gbsn}被盗物品\end{CJK*}, totaling five entities.

1) Classification:

Please determine whether the following case belongs to criminal or civil cases based on the title or relevant description text, and your response should be one of the two options. 

Text: "The People's Procuratorate of Neixiang County accuses that, from June 26 to June 29, 2016, the defendant, Zhang, organized personnel to cut down a large number of poplar trees on the roadside farmland in Shangwangzhuangzu, Miaobei Village, Chimei Town, without obtaining a timber harvesting permit. According to the appraisal by the Neixiang County Forestry Investigation and Design Team, a total of 128 poplar trees were felled, with a total living wood volume of 32.5521 cubic meters. On June 29, 2016, the defendant, Zhang, voluntarily surrendered to the Neixiang County Forest Public Security Bureau."

Answer:

\begin{CJK*}{UTF8}{gbsn}请根据以下案件的标题或者相关描述文本，判断该案件属于刑事案件还是民事案件，并且你的回答应该只能是其中一个。

文本：'内乡县人民检察院指控，2016年6月26日至6月29日期间，被告人张某在未办理林木采伐许可证的情况下，组织人员将其购买的位于赤眉镇庙北村上王庄组路边耕地里的大量杨树砍伐。经内乡县林业调查设计队鉴定，共砍伐杨树128株，计活立木蓄积32.5521立方米。2016年6月29日，被告人张某主动到内乡县森林公安局投案自首。' 

回答:\end{CJK*}

2) Named Entity Recognized

Your task is to extract the '\{entity\}' entity from the text below. If this entity does not exist, the answer is 'No'.

Text: "A set of stolen 'Jingqiu' brand batteries worth 1488 yuan."

Answer: 

{\begin{CJK*}{UTF8}{gbsn} 
你的任务是从下面的文本中提取'\{entity\}'实体，如果不存在这个实体，则回答'No'。

文本：被盗“京球”牌蓄电池一组价值人民币1488元。

回答：

\end{CJK*}}

3) Text Generation

If you are a lawyer, please answer the legal consultation question below based on the real scenario.

'Question': I was driving straight ahead, and a tricycle coming from the opposite direction hit me as it came out of the gas station, causing injuries to people. Who is more responsible, and how is the responsibility divided?

\begin{CJK*}{UTF8}{gbsn}
假设你是一名律师，请回答下面这个真实情景下的中文法律咨询问题。

'问题'：我开车直行，对面三轮车从加油站出来撞了，人受伤了，是谁的责任大，怎么划分责任的?
\end{CJK*}


\subsection{Manual Evaluation Dataset}
\label{appendix:dataset manual}

In this section, we focus on the manual evaluation of the Judicial Reasoning Generation and Legal Consultation tasks.

\textbf{Legal Consultation.} We directly use the legal evaluation dataset from the previous automatic evaluation of the Legal Consultation task, sampling 50 data points as the artificial evaluation dataset for the Legal Consultation task. 

\textbf{Judicial Reasoning Generation.} We reconstructed the evaluation dataset. Our dataset is sourced from the China Judgements Online (CJO), where all are written judgment of first instance. We extract the sections in the documents related to the  court identified that, claims, and court hold that. In the end, our reconstructed Judicial Reasoning Generation manual evaluation dataset consists of 50 data points, covering five charges: kidnapping, trafficking of women and children, fraud, robbery, and extortion, with 10 data points for each charge.

\begin{table*}[!htb]
  \centering
  
  \resizebox{0.999\textwidth}{!}{
  \begin{tabular}{cccccccccccc}
    \toprule
    Capability & Task & Metrics  & GPT-4 & ChatGPT & HanFei & wisdomInterrogatory &  Fuzi-Mingcha & LexiLaw & LaWGPT & Lawyer-LLaMA & ChatLaw \\
    \midrule
    
    \multirow{12}[0]{*}{BIR}  
        & \multirow{4}[0]{*}{Legal Article Recommendation}   &  Acc  & \textbf{0.9890}  & \underline{0.9880}  & 0.1690  & 0.0020  & 0.5540  & 0.5240  & 0.0590  & 0.1280  & 0.6570   \\
        &   & Miss  & 0.0060  & 0.0050  & 0.6530  & 0.9940  & 0.1840  & 0.0100  & 0.8770 & 0.7570 & 0.1000  \\
        &   & F1   & \textbf{0.9920} & \underline{0.9905} & 0.2491 & 0.0039 & 0.5895 & 0.4716 & 0.1015 & 0.2026 & 0.6708  \\
        &   & Mcc  & 0.9832 & 0.9816 & 0.0827 & -0.0054 & 0.3987 & 0.2691 & 0.0293 & 0.0783 & 0.4966  \\
        \cmidrule{2-12}
        & \multirow{4}[0]{*}{Element Recognition}   &  Acc  &  \textbf{0.8170} & \underline{0.7910} & 0.0600 & 0.0010 & 0.1390 & 0.0230 & 0.0480 & 0.0080 & 0.3050 \\
        &   & Miss  &  0 & 0.0010 & 0.7650 & 0.9970 & 0.0750 & 0.8250 & 0.2900 & 0.9700 & 0.2880 \\
        &   & F1   &  \textbf{0.8227} & \underline{0.7932} & 0.0725 & 0.0019 & 0.1258 & 0.0289 & 0.0259 & 0.0152 & 0.3129  \\
        &   & Mcc  &  0.7960 & 0.7656 & 0.0289 & 0.0110 & 0.0861 & 0.0113 & -0.0108 & 0.0198 & 0.2381  \\
        \cmidrule{2-12}
        & Named Entity Recognition  &  Entity-Acc  &  \textbf{0.8067} & \underline{0.6173} & 0.5163 & 0 & 0.0038 & 0.3135 & 0 & 0.0788 & 0.5221 \\
        \cmidrule{2-12}
        & \multirow{3}[0]{*}{Judicial Summarization}   &  ROUGE-1  &  0.5549 & 0.5463 & 0.2834 & 0.4592 & \underline{0.6243} & 0.5406 & 0.3894 & \textbf{0.6467} & 0.5362  \\
        &   & ROUGE-2  &  0.2982 & 0.2849 & 0.1359 & 0.2400 & \underline{0.3423} & 0.2947 & 0.1746 & \textbf{0.3877} & 0.3000   \\
        &   & ROUGE-L  &  0.4285 & 0.3990 & 0.2150 & 0.3433 & \underline{0.4710} & 0.4184 & 0.2668 & \textbf{0.4994} & 0.4036   \\
        \cmidrule{2-12}
        & \multirow{4}[0]{*}{Case Recognition}   &  Acc  &  \textbf{0.9975} & \underline{0.9885} & 0.8270 & 0.2820 & 0.7935 & 0.8380 & 0.4670 & 0.7505 & 0.9815  \\
        &   & Miss  &  0 & 0 & 0 & 0.4435 & 0.0025 & 0.0010 & 0.1790 & 0.0005 & 0.0010   \\
        &   & F1   &  \textbf{0.9975} & \underline{0.9885} & 0.8218 & 0.2799 & 0.7857 & 0.8343 & 0.3692 & 0.7344 & 0.9820   \\
        &   & Mcc  &  0.9950 & 0.9771 & 0.6967 & 0.0085 & 0.6436 & 0.7127 & 0.1451 & 0.5777 & 0.9634    \\
        \cmidrule{1-12}

    \multirow{24}[0]{*}{LFI}  
        & \multirow{4}[0]{*}{Controversy Focus Mining}   &  Acc  &  \textbf{0.8072} & \underline{0.5458} & 0.0229 & 0.0817 & 0.049 & 0.0359 & 0.0458 & 0.0392 & 0   \\
        &   & Miss  &  0.0196 & 0.0196 & 0.3595 & 0.2484 & 0.4085 & 0.6536 & 0.4641 & 0.4967 & 1    \\
        &   & F1   &  \textbf{0.8050} & \underline{0.5716} & 0.0115 & 0.0357 & 0.0470 & 0.0211 & 0.0162 & 0.0219 & 0  \\
        &   & Mcc  & 0.7662 & 0.4713 & -0.0284 & 0.0393 & 0.0066 & 0.0210 & 0.0159 & 0.0079 & 0   \\
        \cmidrule{2-12}
        & \multirow{4}[0]{*}{Similar Case Matching}   &  Acc  &  \textbf{0.5692} & \underline{0.5500} & 0 & 0.3885 & 0.1654 & 0.1231 & 0 & 0.0038 & 0  \\
        &   & Miss  &  0 & 0.0038 & 0.9962 & 0.3423 & 0.6692 & 0.7769 & 1 & 0.9923 & 1  \\
        &   & F1   &  \underline{0.4594} & \textbf{0.4617} & 0 & 0.3538 & 0.2084 & 0.1849 & 0 & 0.0076 & 0    \\
        &   & Mcc  &  0.1078 & 0.0392 & -0.0343 & 0.0566 & -0.0335 & 0.0055 & 0 & 0 & 0   \\
        \cmidrule{2-12}
        & \multirow{4}[0]{*}{Charge Prediction}   &  Acc  &  \textbf{1} & \underline{}{0.9927} & 0.1717 & 0.0121 & 0.2044 & 0.0181 & 0.1330 & 0.0012 & 0.4631   \\
        &   & Miss  & 0 & 0 & 0.0060 & 0.9649 & 0.7352 & 0.9528 & 0.7509 & 0.9915 & 0.0278    \\
        &   & F1   &  \textbf{1} & \underline{0.9928} & 0.0527 & 0.0232 & 0.3153 & 0.0340 & 0.2004 & 0.0024 & 0.3782  \\
        &   & Mcc  &  1 & 0.9880 & 0.0017 & 0.0090 & 0.2135 & 0.0118 & 0.0333 & -0.0186 & 0.0112    \\
        \cmidrule{2-12}
        & \multirow{4}[0]{*}{Prison Term Prediction}   &  Acc  &  \textbf{0.6533} & \underline{0.4499} & 0.0802 & 0.0287 & 0.4097 & 0.0716 & 0.0745 & 0.0115 & 0.2579   \\
        &   & Miss  &  0 & 0 & 0 & 0.7450 & 0.2923 & 0.4900 & 0 & 0.9628 & 0.0573   \\
        &   & F1   &  \textbf{0.6558} & 0.4735 & 0.0273 & 0.0130 & \underline{0.484} & 0.0642 & 0.0103 & 0.0212 & 0.3085   \\
        &   & Mcc  &  0.3353 & 0.1705 & -0.0125 & 0.0239 & 0.0810 & -0.0226 & 0 & 0.0240 & -0.0467   \\
        \cmidrule{2-12}
        & \multirow{4}[0]{*}{Civil Trial Prediction}   &  Acc  &  \underline{0.6775} & 0.5925 & \textbf{0.7675} & 0.0950 & 0.2183 & 0.0266 & 0.5038 & 0.0712 & 0.1500   \\
        &   & Miss  &  0.0525 & 0.0075 & 0.0025 & 0.8950 & 0.6713 & 0.9686 & 0.3425 & 0.8988 & 0.1138  \\
        &   & F1   &  \textbf{0.7043} & 0.6285 & \underline{0.6681} & 0.1676 & 0.3266 & 0.0435 & 0.5455 & 0.1275 & 0.0658   \\
        &   & Mcc  &  0.2657 & 0.1929 & 0.0155 & 0.0602 & 0.0165 & -0.0046 & 0.0023 & 0.0051 & 0.0283   \\
        \cmidrule{2-12}
        & \multirow{4}[0]{*}{Legal Question Answering}   &  Acc  &  \textbf{0.5298} & \underline{0.3789} & 0.2398 & 0.0222 & 0.2456 & 0.2199 & 0.1731 & 0.2175 & 0    \\
        &   & Miss  & 0.0012 & 0 & 0.0538 & 0.8760 & 0.1871 & 0.0959 & 0.2094 & 0.2094 & 1    \\
        &   & F1   &  \textbf{0.5314} & \underline{0.3708} & 0.2203 & 0.0334 & 0.2664 & 0.1851 & 0.0840 & 0.2026 & 0   \\
        &   & Mcc  &  0.3762 & 0.1710 & 0.0284 & -0.0096 & 0.0629 & 0.0199 & 0.0240 & 0.0544 & 0   \\
        \cmidrule{1-12}

    \multirow{9}[0]{*}{CLA}  
        & \multirow{4}[0]{*}{Judicial Reasoning Generation}   &  ROUGE-1  &  0.5193 & 0.4985 & \textbf{0.6882} & 0.2105 & \underline{0.6804} & 0.3613 & 0.4943 & 0.4809 & -   \\
        &   & ROUGE-2  &  0.2473 & 0.238 & \textbf{0.3723} & 0.0698 & \underline{0.3411} & 0.1517 & 0.2286 & 0.2091 & -    \\
        &   & ROUGE-L   &  0.3499 & 0.3326 & \textbf{0.4788} & 0.1371 & \underline{0.4651} & 0.2626 & 0.3340 & 0.3300 & -   \\
        \cmidrule{2-12}
        & \multirow{4}[0]{*}{Case Understanding}   &  ROUGE-1  &  \textbf{0.9650} & \underline{0.9168} & 0.8219 & 0.7502 & 0.8173 & 0.8307 & 0.7187 & 0.8765 & 0.2061    \\
        &   & ROUGE-2  &  \textbf{0.9568} & \underline{0.8919} & 0.7917 & 0.5778 & 0.7837 & 0.7735 & 0.5625 & 0.8268 & 0.1962   \\
        &   & ROUGE-L   &  \textbf{0.9640} & \underline{0.9122} & 0.8220 & 0.7127 & 0.8134 & 0.8200 & 0.6873 & 0.8671 & 0.2047   \\
        \cmidrule{2-12}
        & \multirow{4}[0]{*}{Legal Consultation}   &  ROUGE-1  &  \underline{0.5974} & \textbf{0.6482} & 0.3777 & 0.2518 & 0.4797 & 0.3436 & 0.1956 & 0.4514 & -  \\
        &   & ROUGE-2  &  \underline{0.2758} & \textbf{0.3197} & 0.1693 & 0.0980 & 0.2086 & 0.1391 & 0.0660 & 0.1992 & -   \\
        &   & ROUGE-L   &  \underline{0.4066} & \textbf{0.4585} & 0.2759 & 0.1953 & 0.3346 & 0.2529 & 0.1617 & 0.3044 & -   \\
        
    \bottomrule
    \end{tabular}
}
\caption{The automatic evaluation results of 7 Legal LLMs, GPT-4 and ChatGPT.}
\label{app-tab:legal}
\end{table*}%

\begin{table*}[!htb]
  \centering

  \resizebox{0.999\textwidth}{!}{
  \begin{tabular}{cccccccccccc}
    \toprule
    Capability & Task & Metrics  & Baichuan2-Chat & Baichuan & ChatGLM & Llama-7B &  Llama-13B & Llama2-Chat & Chinese-LLaMA-7B & Chinese-LLaMA-13B & Ziya-LLaMA  \\
    \midrule
    
    \multirow{12}[0]{*}{BIR}  
        & \multirow{4}[0]{*}{Legal Article Recommendation}   &  Acc  & 0.5620 & 0.1800 & 0.7320 & 0.1750 & 0.2660 & 0.4800 & 0.3790 & 0.3580 & 0.6540 \\
        &   & Miss  & 0.0020 & 0.5770 & 0.0030 & 0.6670 & 0.2770 & 0.0170 & 0.0470 & 0.0470 & 0.0020 \\
        &   & F1   & 0.4507 & 0.1781 & 0.7255 & 0.1953 & 0.2816 & 0.4824 & 0.2439 & 0.3034 & 0.6639 \\
        &   & Mcc  & 0.3981 & 0.0097 & 0.6281 & 0.0710 & 0.0202 & 0.2536 & -0.0357 & -0.0344 & 0.4756 \\
        \cmidrule{2-12}
        & \multirow{4}[0]{*}{Element Recognition}   &  Acc  & 0.5400 & 0.0330 & 0.4900 & 0.0370 & 0.1870 & 0.1420 & 0.1310 & 0.0300 & 0.5930 \\
        &   & Miss  & 0 & 0.6200 & 0.0110 & 0.5250 & 0.0240 & 0 & 0.0250 & 0.9080 & 0 \\
        &   & F1   & 0.5218 & 0.0287 & 0.4982 & 0.0143 & 0.0766 & 0.1193 & 0.0745 & 0.0547 & 0.5842 \\
        &   & Mcc  & 0.4995 & -0.0629 & 0.4511 & 0.0054 & -0.0017 & 0.0872 & 0.0293 & 0.0521 & 0.5427 \\
        \cmidrule{2-12}
        & Named Entity Recognition   &  Entity-Acc  & 0.4731 & 0 & 0.0106 & 0 & 0 & 0.0019 & 0 & 0 & 0.4894 \\
        \cmidrule{2-12}
        & \multirow{3}[0]{*}{Judicial Summarization}   &  ROUGE-1  & 0.3584 & 0.3911 & 0.5613 & 0.1655 & 0.1388 & 0.2098 & 0.4094 & 0.1259 & 0.5115 \\
        &   & ROUGE-2  & 0.1632 & 0.1650 & 0.2994 & 0.0584 & 0.0524 & 0.1063 & 0.2174 & 0.0236 & 0.2738 \\
        &   & ROUGE-L  & 0.2785 & 0.2507 & 0.4253 & 0.1180 & 0.1071 & 0.1575 & 0.2963 & 0.0824 & 0.3803 \\
        \cmidrule{2-12}
        & \multirow{4}[0]{*}{Case Recognition}   &  Acc  & 0.9700 & 0.6380 & 0.8735 & 0.2235 & 0.5290 & 0.8360 & 0.5235 & 0.6430 & 0.9470 \\
        &   & Miss  & 0.0030 & 0 & 0.0940 & 0.5130 & 0.0395 & 0 & 0.1450 & 0 & 0.0010 \\
        &   & F1   & 0.9714 & 0.5845 & 0.9127 & 0.2323 & 0.4680 & 0.8317 & 0.4897 & 0.6156 & 0.9473 \\
        &   & Mcc  & 0.9409 & 0.3986 & 0.7876 & -0.0389 & 0.1326 & 0.7095 & 0.2219 & 0.3395 & 0.8988 \\
        \cmidrule{1-12}

    \multirow{24}[0]{*}{LFI}  
        & \multirow{4}[0]{*}{Controversy Focus Mining}   &  Acc  & 0.0621 & 0.0556 & 0.0948 & 0.0425 & 0.0588 & 0.0098 & 0.0229 & 0.0621 & 0.0915 \\
        &   & Miss  & 0.2941 & 0.1405 & 0.7092 & 0.183 & 0.2059 & 0.6863 & 0.6373 & 0.1732 & 0.0327 \\
        &   & F1   & 0.0412 & 0.0174 & 0.1418 & 0.0131 & 0.0186 & 0.0074 & 0.0202 & 0.0328 & 0.0564 \\
        &   & Mcc  & 0.0186 & -0.0061 & 0.1105 & -0.0198 & 0.0059 & -0.0206 & -0.0020 & 0.0069 & 0.0052 \\
        \cmidrule{2-12}
        & \multirow{4}[0]{*}{Similar Case Matching}   &  Acc  & 0.0154 & 0 & 0.5500 & 0 & 0 & 0 & 0.0038 & 0.0269 & 0.0038 \\
        &   & Miss  & 0.9692 & 1 & 0 & 1 & 1 & 1 & 0.9962 & 0.9538 & 0.9962 \\
        &   & F1   & 0.0299 & 0 & 0.3903 & 0 & 0 & 0 & 0.0076 & 0.0505 & 0.0076 \\
        &   & Mcc  & -0.0045 & 0 & 0 & 0 & 0 & 0 & 0.0343 & 0.0092 & 0.0281 \\
        \cmidrule{2-12}
        & \multirow{4}[0]{*}{Charge Prediction}   &  Acc  & 0.2406 & 0.0060 & 0.6010 & 0.4317 & 0.4643 & 0.3857 & 0.3362 & 0.1391 & 0.5998 \\
        &   & Miss  & 0 & 0.9794 & 0.2902 & 0.2273 & 0.1016 & 0.2648 & 0.3277 & 0.6784 & 0.0073 \\
        &   & F1   & 0.1750 & 0.0118 & 0.6757 & 0.3519 & 0.3679 & 0.3879 & 0.3179 & 0.2021 & 0.5318 \\
        &   & Mcc  & 0.1554 & -0.0011 & 0.5686 & 0.0474 & 0.006 & 0.077 & -0.0197 & 0.0038 & 0.3043 \\
        \cmidrule{2-12}
        & \multirow{4}[0]{*}{Prison Term Prediction}   &  Acc  & 0.7249 & 0.0745 & 0.4155 & 0.0229 & 0.0458 & 0.0860 & 0.0745 & 0.1003 & 0.5616 \\
        &   & Miss  & 0 & 0 & 0.0630 & 0.7393 & 0.6762 & 0.1232 & 0 & 0 & 0 \\
        &   & F1   & 0.6143 & 0.0103 & 0.4484 & 0.0103 & 0.0580 & 0.0731 & 0.0103 & 0.0533 & 0.5562 \\
        &   & Mcc  & 0.0533 & 0 & 0.0871 & 0.0040 & 0.0096 & -0.0347 & 0 & 0.0539 & -0.0377 \\
        \cmidrule{2-12}
        & \multirow{4}[0]{*}{Civil Trial Prediction}   &  Acc  & 0.6875 & 0.7037 & 0.2334 & 0.4200 & 0.3063 & 0.5750 & 0.7262 & 0.7113 & 0.2787 \\
        &   & Miss  & 0.0013 & 0.0875 & 0.6512 & 0.4537 & 0.6050 & 0.1562 & 0.0525 & 0.0525 & 0.0063 \\
        &   & F1   & 0.6791 & 0.6450 & 0.3302 & 0.4915 & 0.4046 & 0.6209 & 0.6524 & 0.6446 & 0.3607 \\
        &   & Mcc  & 0.1544 & 0.0196 & -0.0403 & 0.0022 & 0.0061 & 0.1081 & -0.0064 & -0.0275 & -0.0348 \\
        \cmidrule{2-12}
        & \multirow{4}[0]{*}{Legal Question Answering}   &  Acc  & 0.3836 & 0.2304 & 0.2491 & 0.1193 & 0.0772 & 0.0164 & 0.1591 & 0.1497 & 0.2608 \\
        &   & Miss  & 0.0152 & 0.1368 & 0.0234 & 0.3519 & 0.6386 & 0.9404 & 0.2070 & 0.3988 & 0.0012 \\
        &   & F1   & 0.3824 & 0.2432 & 0.2386 & 0.0574 & 0.0557 & 0.0259 & 0.0863 & 0.1660 & 0.2538 \\
        &   & Mcc  & 0.1850 & 0.0178 & 0.0347 & -0.0170 & 0.0070 & 0.0147 & -0.0064 & 0.0005 & 0.0361 \\
        \cmidrule{1-12}

    \multirow{9}[0]{*}{CLA}  
        & \multirow{4}[0]{*}{Judicial Reasoning Generation}   &  ROUGE-1  & 0.6967 & 0.5295 & 0.5096 & 0.0088 & 0.1663 & 0.4052 & 0.3692 & 0.2602 & 0.4113 \\
        &   & ROUGE-2  & 0.3938 & 0.2974 & 0.2158 & 0.0033 & 0.0616 & 0.1759 & 0.1633 & 0.1053 & 0.1948 \\
        &   & ROUGE-L  & 0.4878 & 0.3811 & 0.3363 & 0.0062 & 0.1077 & 0.2816 & 0.2578 & 0.2004 & 0.2975 \\
        \cmidrule{2-12}
        & \multirow{4}[0]{*}{Case Understanding}   &  ROUGE-1  & 0.8249 & 0.3857 & 0.8821 & 0.5995 & 0.7009 & 0.7175 & 0.6745 & 0.7718 & 0.8562 \\
        &   & ROUGE-2  & 0.7920 & 0.2574 & 0.8480 & 0.4948 & 0.5912 & 0.6584 & 0.5441 & 0.6717 & 0.8150 \\
        &   & ROUGE-L  & 0.8219 & 0.3707 & 0.8769 & 0.5880 & 0.6784 & 0.7093 & 0.6507 & 0.7510 & 0.8477 \\
        \cmidrule{2-12}
        & \multirow{4}[0]{*}{Legal Consultation}   &  ROUGE-1  & 0.5882 & 0.2508 & 0.5007 & 0.1496 & 0.1555 & 0.2618 & 0.1912 & 0.1699 & 0.3494 \\
        &   & ROUGE-2  & 0.2547 & 0.0973 & 0.2022 & 0.0500 & 0.0505 & 0.0885 & 0.0664 & 0.0586 & 0.1529 \\
        &   & ROUGE-L  & 0.3963 & 0.2071 & 0.3478 & 0.1283 & 0.1343 & 0.1793 & 0.1568 & 0.1434 & 0.2554 \\

    \bottomrule
    \end{tabular}
}
  \caption{The automatic evaluation results of baseline LLMs.}
  \label{app-tab:baseline}
\end{table*}%

\begin{table}[!htb]
  \centering

  \resizebox{0.4999\textwidth}{!}{
  \begin{tabular}{c|ccc|ccc}
    \toprule
    
    \multirow{2}[0]{*}{Model} & \multicolumn{3}{c|}{Judicial Reasoning Generation}  & \multicolumn{3}{c}{Legal Consultation}  \\
    \cmidrule{2-7}
     & $WR_{A}$  & $WR_{B}$  & $WR_{C}$  & $WR_{A}$  & $WR_{B}$  & $WR_{C}$ \\
    \midrule
    GPT-4 & 0.34  & 0.22  & 0.58  & 0.98  & 0.88  & 0.68  \\
    \cmidrule{1-7}
    ChatGPT & 0.22  & 0.18 & 0.66 & 0.82  & 0.90  & 0.66  \\
    \cmidrule{1-7}
    Fuzi-Mingcha & 0.74  & 0.26  & 0.94  & 0.40  & 0.72  & 0.40 \\
    \cmidrule{1-7}
    HanFei & 0.58  & 0.34  & 0.86  & 0.34  & 0.38  & 0.26  \\
    \cmidrule{1-7}
    LexiLaw & 0.18 & 0.28  & 0.48  & 0.22 & 0.26  & 0.24 \\
    \cmidrule{1-7}
    Lawyer-LLaMA & 0.18 & 0.12  & 0.60  & 0.46  & 0.74  & 0.32  \\
        
    \bottomrule
  \end{tabular}
}
    \caption{The win rate (WR) of LLMs for the Judicial Reasoning Generation and Legal Consultation tasks. Subscripts A, B, C represent the judgment results of three experts respectively.}
  \label{tab:win rate all}%
\end{table}%

\section{Data Annotation}
\subsection{Data License}

The Legal Consultation annotation data is sourced from a public dataset, while the Judicial Reasoning Generation annotation data comes from our private dataset. All personally identifiable information such as names, phone numbers, and ID numbers has been anonymized in the process. Therefore, we can proceed with annotating these two datasets.

\subsection{Annotation Rules and Standards}
\label{Annotation rules and standards}

\begin{figure}[!htb]
\centering
\includegraphics[width=1\columnwidth]{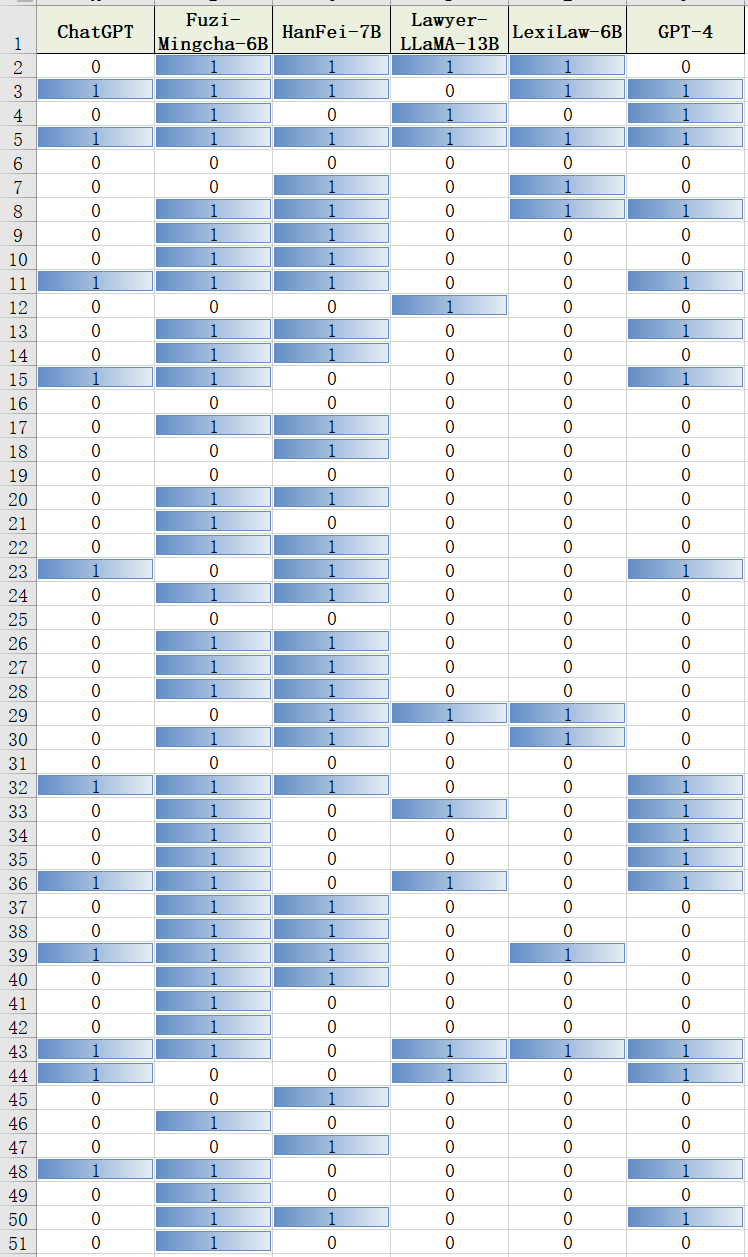} 
\caption{The annotation results of expert A for the Judicial Reasoning Generation task. And this annotation is based on using the reference answer as the baseline.}
\label{fig:Expert-A-tag}
\end{figure}

Before starting the annotation process, we formulated annotation guidelines for the Judicial Reasoning Generation and Legal Consultation tasks through discussions with legal experts. 

For the Judicial Reasoning Generation task, the criteria are completeness, relevance and accuracy.
\begin{itemize}
    \item Completeness: Whether the reasoning content is complete, including the completeness of the reasoning structure and whether explicit penalties are provided.
    \item Relevance: The degree of relevance between the reasoning content and the case.
    \item Accuracy: Whether the reasoning content is accurate, including the presence of fabricated facts, incorrect citation of legal provisions, and usage errors.
\end{itemize}

As for the Legal Consultation task, the criteria include flueny, relevance and comprehensibility.
\begin{itemize}
    \item Fluency: The fluency and coherence of the response content.
    \item Relevance: The relevance of the response content to legal issues and its alignment with legal practicality.
    \item Comprehensibility: The level of understanding of legal issues in the response content.
\end{itemize}

Additionally, to facilitate computer processing, we standardized the annotation rules for legal experts. For each sample, if the output of the target LLM is better than the baseline, it is marked as 1; otherwise, it is marked as 0. 

During the annotation process, we imported the annotated data into Excel.  Each row represents the input for one data point and the outputs of different models. To prevent potential subjective biases from experts toward LLMs, we adopted a model-anonymous annotation approach. Specifically, for each row, we shuffled the order of models, and the shuffling results varied, ensuring that experts wouldn't know which LLM produced the output during annotation.

Finally, we organized the expert annotations to calculate the win rate for each LLM. Figure \ref{fig:Expert-A-tag} illustrates the annotation results of expert A for the Judicial Reasoning Generation task.

\subsection{Risk Statement}
This work is solely intended for academic research and strictly prohibited for any other commercial activities. Before the annotation process, due to the sensitivity of the legal field, we confirmed the usability and security of the dataset and legal experts have conducted ethical evaluations. Additionally, legal experts have conducted ethical evaluations.

\subsection{Annotators}
The three legal experts conducting the annotations are three graduate students from our research team, specializing in the field of criminal law.

\section{More Results}
\subsection{The Automatic Evaluation Results}
\label{appendix:results of LLMs evaluation}
As shown in Table \ref{app-tab:legal} and Table \ref{app-tab:baseline}, we can observe that their performance is consistent with the trend of our score results. GPT-4 and ChatGPT have strong multi-level capabilities, with a certain legal logic, while other LLMs have strong text generation capabilities but lack legal logic. 

These detailed tables can also help us more clearly identify the strengths and weaknesses of LLMs in various tasks.
The legal LLMs performed unsatisfactorily in tasks corresponding to the major and minor premises in syllogism, such as Legal Article Recommendation and Element Recognition. 
They also fell short in further reasoning tasks such as Charge Prediction, Prison Term Prediction, and Civil Trial Prediction compared to GPT-4 and ChatGPT. 
Overall, the performance of these LLMs indicates a lack of information retrieval and reasoning related to legal logic.

\subsection{The Win Rate of LLMs for Each Expert}
\label{appendix:win rate for each expert}

As shown in Table \ref{tab:win rate all}, Expert A and B have similar win rates, while Expert C differs significantly from them. This suggests that while legal logic is commonly recognized among legal experts, there are still individual differences in actual judgment, influenced by certain subjectivity.

\subsection{The Agreement Scores for Expert Evaluation}
\label{The agreement scores}

\begin{table}[!htb]
  \centering

  \resizebox{0.35\textwidth}{!}{
  \begin{tabular}{ccc}
    \toprule
    
    Model & $JRG_{ref}$ & $LC_{ref}$  \\
    \midrule
    GPT-4 & 0.57 & 0.77 \\
    \cmidrule{1-3}
    ChatGPT & 0.55 & 0.69  \\
    \cmidrule{1-3}
    Fuzi-Mingcha & 0.52 & 0.59  \\
    \cmidrule{1-3}
    HanFei & 0.55 & 0.71 \\
    \cmidrule{1-3}
    LexiLaw & \textbf{0.63} & \textbf{0.80}  \\
    \cmidrule{1-3}
    Lawyer-LLaMA & 0.53 & 0.52  \\
        
    \bottomrule
    \end{tabular}
}
    \caption{The agreement scores of LLMs. JRG and LC represent the Judicial Reasoning Generation and Legal Consultation tasks, respectively. The subscript $_{ref}$ indicates the agreement of the evaluations from the three experts when using the reference answer as the baseline. }
  \label{tab:agreement score}%
\end{table}%

Furthermore, for the manual evaluation, we calculated agreement scores for expert evaluation, as shown in Table \ref{tab:agreement score}. Based on this, we observe the following fact:

\textbf{Although experts can find the lack of legal logic in LLMs, assessing legal logic may also pose a challenge for experts.}
The agreement score for the Judicial Reasoning Generation task is noticeably lower than that for the Legal Consultation task. 
The reference answers for judicial reasoning generation tasks are derived from actual court judgments in legal documents, serving as the gold answers. This task emphasizes the completeness and accuracy of formal content, which is directly related to legal logic. 
This allows experts to judge based on their legal logic, which may be affected by their legal background, bring noise, and also bring challenges to evaluation.

On the other hand, legal consultation work involves legal opinions for the public, covering a broader range of legal areas but addressing common legal issues. 
Experts provide answers more based on fluency rather than based on the legal logic of legal practice. 
This makes it easier for experts to judge, and the agreement scores are higher.

\section{Responsible NLP Research Checklist}

\subsection{For every submission}

\subsubsection{Did you discuss the \textit{limitations} of your work?}

Yes. Section 7.
        
\subsubsection{Did you discuss any potential \textit{risks} of your work?}

Yes. Section 8.

\subsubsection{Do the abstract and introduction summarize the paper’s main claims?}

Yes. Abstract / Section 1.

\subsection{Did you use or create \textit{scientific artifacts}?}

Yes.

If yes:
\subsubsection{Did you cite the creators of artifacts you used?}

Yes. The creators of these datasets and LLMs are presented in Table 1 of Section 3.2.1, Table 2 of Section 5.1, respectively.

\subsubsection{Did you discuss the \textit{license or terms} for use and/or distribution of any artifacts?}

Yes. As shown in Section 3.2 and Table 2 of Section 5.1.  All datasets we used are publicly available. Regarding the evaluated LLMs, GPT models were accessed through API, Llama-type models required application through the official website, and the rest of the LLMs were open source.

\subsubsection{Did you discuss if your use of existing artifact(s) was consistent with their \textit{intended use}, provided that it was specified? For the artifacts you create, do you specify intended use and whether that is compatible with the original access conditions (in particular, derivatives of data accessed for research purposes should not be used outside of research contexts)?}

No. Because their licenses are well-known, allowing works like ours to use these artifacts.

\subsubsection{Did you discuss the steps taken to check whether the data that was collected/used contains any \textit{information that names or uniquely identifies individual people} or \textit{offensive content}, and the steps taken to protect / anonymize it?}

Yes. Section 8.

\subsubsection{Did you provide documentation of the artifacts, e.g., coverage of domains, languages, and linguistic phenomena, demographic groups represented, etc.?}

Yes. Section 2, Section 3.2,  Appendix A.1 and Appendix A.3. We provided the sources, language types, legal domains, and other information for all datasets used in this work. And The evaluated LLMs cover the existing mainstream Chinese legal LLMs.

\subsubsection{Did you report relevant statistics like the number of examples, details of train/test/dev splits, etc. for the data that you used/created?}

Yes. Table 1 of Section 3.2,  Appendix A.1 and Appendix A.3.

\subsection{Did you run \textit{computational experiments}?} 

Yes.

If yes:
\subsubsection{Did you report the \textit{number of parameters} in the models used, the \textit{total computational budget} (e.g., GPU hours), and \textit{computing infrastructure} used?}

No. Because we aim to construct a LLMs benchmark, we do not involve fine-tuning. However, we provided the sizes of all parameters for each evaluated LLM, as shown in Table 2 of Section 5.1.

\subsubsection{Did you discuss the experimental setup, including \textit{hyperparameter search} and \textit{best-found hyperparameter} values?}

Yes and no. Our work is to construct a legal LLMs benchmark, which does not involve training or fine-tuning LLMs. However, we provide detailed information on the evaluation dataset, models, and other aspects, as shown in Section 5.1.

\subsubsection{Did you report \textit{descriptive statistics} about your results (e.g., error bars around results, summary statistics from sets of experiments), and is it transparent whether you are reporting the max, mean, etc. or just a single run?}

Yes. Section 4. In Section 4.1, we present specific metrics for automatic evaluation, all experimental results, encompassing all LLMs and metrics mentioned in this paper, are provided in Appendix C.1. In Section 4.2, we present metrics for manual evaluation, with a comprehensive set of results detailed in Table 4 and the complete evaluation results available in Appendix C.2 and Appendix D.

\subsubsection{If you used existing packages (e.g., for preprocessing, for normalization, or for evaluation), did you report the implementation, model, and parameter settings used (e.g., NLTK, Spacy, ROUGE, etc.)?}

No. The project utilized a variety of packages, and in the end, we open-sourced all the code along with the corresponding runtime environment.

\subsection{Did you use \textit{human annotators} (e.g., crowdworkers) or \textit{research with human subjects}?}  

Yes

If yes:
\subsubsection{Did you report the full text of instructions given to participants, including e.g., screenshots, disclaimers of any risks to participants or annotators, etc.?}

Yes. Appendix B.2 and Appendix B.3.

\subsubsection{Did you report information about how you recruited (e.g., crowdsourcing platform, students) and paid participants, and discuss if such \textit{payment is adequate} given the participants’ demographic (e.g., country of residence)?}

Yes. Appendix B.4. The annotators are law graduate students from our research team, and their remuneration will be uniformly covered by the project team.

\subsubsection{Did you discuss whether and how \textit{consent} was obtained from people whose data you're using/curating (e.g., did your instructions explain how the data would be used)?}

Yes. Appendix B.1.

\subsubsection{Was the data collection protocol \textit{approved (or determined exempt)} by an ethics review board?}

Yes. Appendix B.1 and Appendix B.3.  This is a reannotation of openly available online data.

\subsubsection{Did you report the basic demographic and geographic characteristics of the \textit{annotator} population that is the source of the data?}

Yes. Appendix B.4.

\subsection{Did you use \textit{AI assistants} (e.g., ChatGPT, Copilot) in your research, coding, or writing?}

Yes.

\subsubsection{Did you include information about your use of AI assistants?}

Yes. Section 5.1.

\end{document}